\documentclass{article}

\PassOptionsToPackage{numbers, compress}{natbib}
\usepackage{titletoc}
\newcommand\DoToC{%
  \startcontents
  \printcontents{}{1}{\hrulefill\vskip0pt}
  \vskip0pt \noindent\hrulefill
  }

\usepackage[preprint]{neurips_2022}
\usepackage{subfigure}
\usepackage{ulem}
\usepackage[table]{xcolor}
\usepackage{colortbl}
\usepackage{amsmath}
\usepackage{enumitem}
\usepackage{booktabs}
\usepackage{makecell}
\usepackage{algorithm}
\usepackage{amsfonts, amssymb}
\usepackage{algorithmic}
\usepackage{mathtools}
\usepackage{amsthm}
\usepackage[mathscr]{eucal}
\usepackage{bm}
\usepackage{graphicx}
\usepackage{multirow}
\usepackage{amsmath,lipsum}
\usepackage{amssymb} 
\usepackage{wrapfig}
\usepackage{ulem}

\usepackage{tabularx}
\usepackage{cuted}%%\stripsep-3pt
\usepackage[utf8]{inputenc}
\usepackage{ragged2e}

\usepackage[english]{babel}

\newtheorem{theorem}{Theorem}
\newtheorem{corollary}{Corollary}[theorem]

\usepackage{subfiles}
\usepackage{xr}
\definecolor{myblue}{HTML}{b2f0ff}
\definecolor{myblue2}{HTML}{cef5ff}
\definecolor{myblue3}{HTML}{e7faff}
\usepackage{color}

\usepackage{amsmath,amsfonts,bm}
\def\eqref#1{equation~\ref{#1}}
\def\1{\bm{1}}

\def\rvx{{\mathbf{x}}}

\def\rvz{{\mathbf{z}}}

\def\mJ{{\bm{J}}}

\DeclareMathAlphabet{\mathsfit}{\encodingdefault}{\sfdefault}{m}{sl}
\SetMathAlphabet{\mathsfit}{bold}{\encodingdefault}{\sfdefault}{bx}{n}

\usepackage[utf8]{inputenc} % allow utf-8 input
\usepackage[T1]{fontenc}    % use 8-bit T1 fonts
\usepackage{hyperref}       % hyperlinks
\usepackage{url}            % simple URL typesetting
\usepackage{booktabs}       % professional-quality tables
\usepackage{amsfonts}       % blackboard math symbols
\usepackage{nicefrac}       % compact symbols for 1/2, etc.
\usepackage{microtype}      % microtypography
\usepackage{xcolor}         % colors
\title{From Orthogonality to Dependency: Learning Disentangled Representation for Multi-Modal Time-Series Sensing Signals}

 \author{%/
Ruichu Cai$^{2}$,  \textbf{Zhifan Jiang}$^{2}$, \textbf{Zijian  Li}$^{3,}$\thanks{Equal contributions}, Weilin Chen$^{2}$, Xuexin Chen$^{2}$, \\ \textbf{Zhifeng Hao}$^{4}$, \textbf{Yifan Shen}$^{5}$,  \textbf{Guangyi Chen}$^{3,1}$, \textbf{Kun Zhang}$^{3,1}$~\\$^1$ Carnegie Mellon University \\ $^2$ School of Computer Science, Guangdong University of Technology \\$^3$ Mohamed bin Zayed University of Artificial Intelligence\\$^4$ Shantou University 
 }

\begin{document}

\maketitle

% 尽管TDL已经做了很好了，但是依赖xxx假设，最近非常好的努力解决这个问题，但是需要xxxx假设
\begin{abstract}
Existing methods for multi-modal time series representation learning aim to disentangle the modality-shared and modality-specific latent variables. Although achieving notable performances on downstream tasks, they usually assume an orthogonal latent space. However, the modality-specific and modality-shared latent variables might be dependent on real-world scenarios. Therefore, we propose a general generation process, where the modality-shared and modality-specific latent variables are dependent, and further develop a \textbf{M}ulti-mod\textbf{A}l \textbf{TE}mporal Disentanglement (\textbf{MATE}) model. Specifically, our \textbf{MATE} model is built on a temporally variational inference architecture with the modality-shared and modality-specific prior networks for the disentanglement of latent variables. Furthermore, we establish identifiability results to show that the extracted representation is disentangled. More specifically, we first achieve the subspace identifiability for modality-shared and modality-specific latent variables by leveraging the pairing of multi-modal data. Then we establish the component-wise identifiability of modality-specific latent variables by employing sufficient changes of historical latent variables. Extensive experimental studies on multi-modal sensors, human activity recognition, and healthcare datasets show a general improvement in different downstream tasks,  highlighting the effectiveness of our method in real-world scenarios. 
% our method significantly surpasses existing time series forecasting techniques. 
\end{abstract}

% 一个名字对应一个东西

\section{Introduction}
% Time series analysis 需要是被隐dynamic
% 现有方法
% 但是不能解决instantaneous
% 我们的方法

% 帮助sora理解物理规则

Most of the existing works for time series analysis \cite{zhang2024self,liang2024foundation,li2023difformer,wu2022timesnet,luo2024moderntcn,zhou2024one} are usually devised for homogeneous data, with the assumption that time series are sampled from the same modality. However, the heterogeneous time series data \cite{shao2023exploring,liu2024mtsa,liu2024focal}, which are sampled from multiple modalities and not compatible with these methods, are also common in several real-world applications, e.g., Internet of Things (IoT) \cite{radu2016towards,xu2016indoor,nonnenmacher2022utilizing}, health care 
\cite{zhang2022m3care,makarious2022multi,iglesias2023ready}, and finance \cite{cheng2022financial,zhou2020domain}. To model the multi-modal time series data, one mainstream solution is to disentangle the modality-specific and modality-shared latent variables from the observational time series signal.

% Therefore, how to model the multi-modal time series data has become a critical challenge.
% To well model the multi-modal time series data, different methods are proposed. 

% 分类：
% 对比学习
% 可识别性

% 1. Focal
% 2，IJCAI
% 3. cocoa
% 4. Cosmo
% To model the multi-modal time series data, one mainstream solution is to disentangle the modality-specific and modality-invariant latent variables from the observations. 

% To disentangle the modality-specific and modality-shared temporally latent variables, several methods are proposed.
Several methods are proposed to disentangle the modality-specific and modality-shared temporally latent variables. One mainstream approach is based on the contrastive learning method. For example, Deldari et.al proposes COCOA \cite{deldari2022cocoa}, which learns modality-shared representations by aligning the representation from the same timestamp, and Ouyang et.al propose Cosmo \cite{ouyang2022cosmo}, which extracts modality-shared representations by using a iterative fusion learning strategy. Considering that the modality-specific representations also play an important role in the downstream task, Liu et.al \cite{liu2024focal} use an orthogonality restriction and simultaneously leverage the modality-shared and modality-specific representations. Considering the multi-view setting as a special case of the multi-modal setting, Huang et.al \cite{huang2023latent} develop the identifiability results of the latent temporal process by minimizing the contrastive objective function. In summary, these methods usually assume that the modality-shared and modality-specific latent variables are orthogonal, hence they can be disentangled by using different contrastive-learning-based constraints. Please refer to Appendix \ref{app:rl} for further discussion of related works, including multi-modal representation learning, multi-modal time series modeling, and the identifiability of generative models.

\begin{wrapfigure}{r}{8.5cm}
    \centering
    \vspace{-2.5mm}
    \includegraphics[width=0.6\columnwidth]{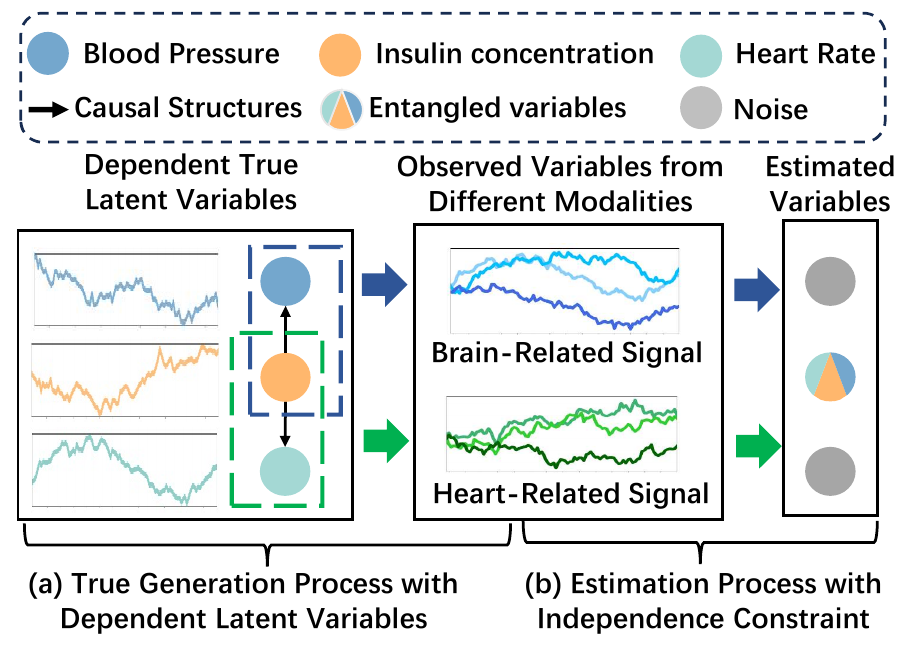}
    % \vspace{-1mm}
    \caption{Illustration of physiological indicators of diabetics, where brain-related and heart-related signals are observations. (a) In the true generation process, observations are generated from dependent latent sources. (b) In the estimation process, enforcing orthogonality on estimated sources can result in the entanglement of latent sources and meaningless noises.}
    \vspace{-1mm}
    \label{fig:dependent_example}
\end{wrapfigure}
Although these methods achieve outstanding performance on several applications, the orthogonality of modality-shared and modality-specific latent space may be too difficult to satisfy in real-world scenarios. Figure \ref{fig:dependent_example} provides an example of physiological indicators of diabetics, where brain-related and heart-related signals are observed in time series data. Specifically, Figure \ref{fig:dependent_example} (a) denotes the true data generation process, where the causal directions from insulin concentration to blood pressure and heart rate denote how diabetes leads to complications of heart disease and high blood pressure. As shown in Figure \ref{fig:dependent_example} (b), existing methods that apply orthogonal constraints on the estimated latent variables despite the dependent true latent sources, lead to the entanglement of latent variables and further the suboptimal performance of downstream tasks.

To address the aforementioned challenge of dependent latent sources, we propose a multi-modal temporal disentanglement framework to estimate the ground-truth latent variables with identifiability guarantees. Specifically, we first leverage the pair-wise multi-modal data to establish the subspace identifiability of latent variables. Sequentially, we leverage the independent influence of historical latent variables to further show the component-wise identifiability of latent variables. Building on the theoretical results, we develop the \textbf{M}ulti-mod\textbf{A}l \textbf{TE}mporal Disentanglement (\textbf{MATE}) model, which incorporates variational inference neural architecture with modality-shared and modality-specific
prior networks. The proposed \textbf{MATE} is validated through extensive downstream tasks for multi-modal time series data. The impressive performance that outperforms state-of-the-art methods demonstrates its effectiveness in real-world applications.
% \section{Related Works}
% \subsection{Identifiability of Latent Variables}

% \subsection{Instantaneous of Time Series Analysis}

% \subsection{Time Series Forecasting}

% \clearpage
\section{Problem Setup}
\subsection{Data Generation Process of Multi-modal Time Series}
\begin{wrapfigure}{r}{7cm}
    \centering
    \vspace{-4mm}
    \includegraphics[width=0.5\textwidth]{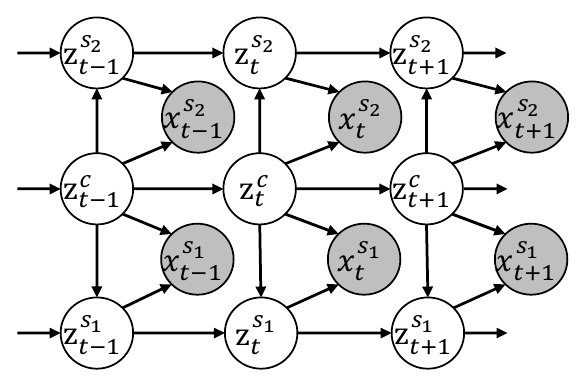}
    \vspace{-5mm}
    \caption{Data generation process of time series data with two modalities. The grey and white nodes denote the observed and latent variables.}
    \label{fig:generation}
    % \vspace{-3mm}
\end{wrapfigure}

To show how to learn disentangled representation for multi-modal time series data, we first introduce the data generation process as shown in Figure \ref{fig:generation}. Specifically, we assume that the existence of $M$ modalities $\mathcal{S}=\{S_1, S_2, \cdots, S_M\}$. For each modality $S_m$, time series data with discrete time steps $\rvx_{1:T}^{s_m}=\{\rvx_1^{s_m}, \rvx_2^{s_m},\cdots,\rvx_T^{s_m}\}$ with the length of $T$ are drawn from a distinct distribution, represented as $p(\rvx_{1:T}^{s_m})$. Moreover, $\rvx_t^{s_m}$ is generated from the modality-shared and modality-specific latent variables $\rvz_t^c, \rvz_t^{s_m}$ by an invertible and nonlinear mixing function $g_m$ shown as follows:
\begin{equation}
\label{equ:g1}
% \small
    \rvx^{s_m}_t=g_m(\rvz_t^c,\rvz_t^{s_m}).
\end{equation}
For convenience, we let $\rvz_t^m=\{\rvz_t^c,\rvz_t^{s_m}\}$ be the latent variables of $m$-th modality. And we further let $\rvz_t^c=(z_{t,i}^c)_{i=1}^{n_c}$ and $\rvz_{t}^{s_m}=(z_{t,i}^{s_m})_{i=n_c+1}^{n}$. More specifically, the $i$-th dimension modality-shared latent variables $z_{t,i}^c$ are time-delayed and related to the historical modality-shared latent variables $\rvz_{t-\tau}^c$ with the time lag of $\tau$ via a nonparametric function $f^c_i$. Similarly, the modality-specific latent variables are generated via another nonparametric function $f_i^m$, which are formalized as follows:
\begin{equation}
\label{equ:g2}
\small
\begin{split}
    z_{t,i}^c=f_{i}^c(\textit{PA}(z_{t,i}^c), \epsilon^c_{t,i}), \quad \epsilon^c_{t,i}\sim p_{\epsilon^c_{t,i}}\quad\quad
    z_{t,i}^{s_m}=f_{i}^{m}(\textit{PA}(z_{t,i}^{s_m}), \epsilon^{s_m}_{t,i}), \quad \epsilon^{s_m}_{t,i}\sim p_{\epsilon^{s_m}_{t,i}},
\end{split}
\end{equation}
where $\textit{PA}$ denote the set of latent variables that directly cause $z_{t,i}^c$ or $z_{t,i}^{s_m}$, and $\epsilon^{s_m}_{t,i}, \epsilon^c_{t,i}$ denote the independent noise. Combining the example of diabetics in Figure \ref{fig:dependent_example}, $\rvx_t^{s_1}$ and $\rvx_t^{s_2}$ can be considered as brain-related and heart-related signals, respectively. The modality-shared variables $\rvz_t^c$ denote the insulin concentration and $\rvz_t^{s_1}, \rvz_t^{s_2}$ denote the blood pressure and heart rate, respectively. $\rvz_t^c \rightarrow \{\rvz_t^{s_1},\rvz_t^{s_2}\}$ denotes that insulin concentration influences blood pressure and heart rate.
\subsection{Problem Definition}
Based on the aforementioned data generation process, we further provide the problem definition. Specifically, We first suppose to have a set of $M$ sensory modalities. Then, for each group of time series from $M$ modalities, we let $y$ be the corresponding label. Given the labeled multi-modal time series training set with the size of $D$, i.e., $\{X_i,y_i\}_{i=1}^{D}$, we aim to obtain a model that can extract disentangled representations for multi-modal time series data, which can benefit the downstream tasks, i.e. estimate correct label. More mathematically, our goal is to estimate the distribution of the modality-specific latent variables $p(\rvz_{1:T}^{s_1}),\cdots,p(\rvz_{1:T}^{s_M})$ and the modality-shared latent variables $p(\rvz_{1:T}^{c})$ by modeling the observed multi-modal time series data, which are formalized as follows:
\begin{equation}
 \small
\begin{split}
    \ln p(\rvx^{s_1}_{1:T},\cdots,\rvx^{s_M}_{1:T})=\int_{\rvz_{1:T}^{s_1}}&\cdots\int_{\rvz_{1:T}^{s_M}}\int_{\rvz_{1:T}^{c}}\Big(\ln p(\rvx^{s_1}_{1:T},\cdots,\rvx^{s_M}_{1:T}|\rvz_{1:T}^{s_1},\cdots,\rvz_{1:T}^{s_M},\rvz_{1:T}^{c}) \\&+ \sum_{m=1}^M \ln p(\rvz_{1:T}^{s_m}|\rvz_{1:T}^{c}) + \ln p(\rvz_{1:T}^{c})\Big)d\rvz_{1:T}^{s_1}\cdots d\rvz_{1:T}^{s_M}d\rvz_{1:T}^{c}.
\end{split}
\end{equation}
Therefore, to achieve this goal, we first devise a temporal variational inference architecture with prior networks to reconstruct the modality-specific and modality-shared latent variables, which are shown in Section \ref{sec:model}. Sequentially, we further propose theoretical analysis to show that these estimated modality-shared and modality-specific latent variables are identifiable, which are shown in Section \ref{sec:the}.
\section{\textbf{MATE:} Multi-modal Temporal Disentanglement Model}\label{sec:model}
\begin{figure}[t]
 \small
    \centering
    \includegraphics[width=\columnwidth]{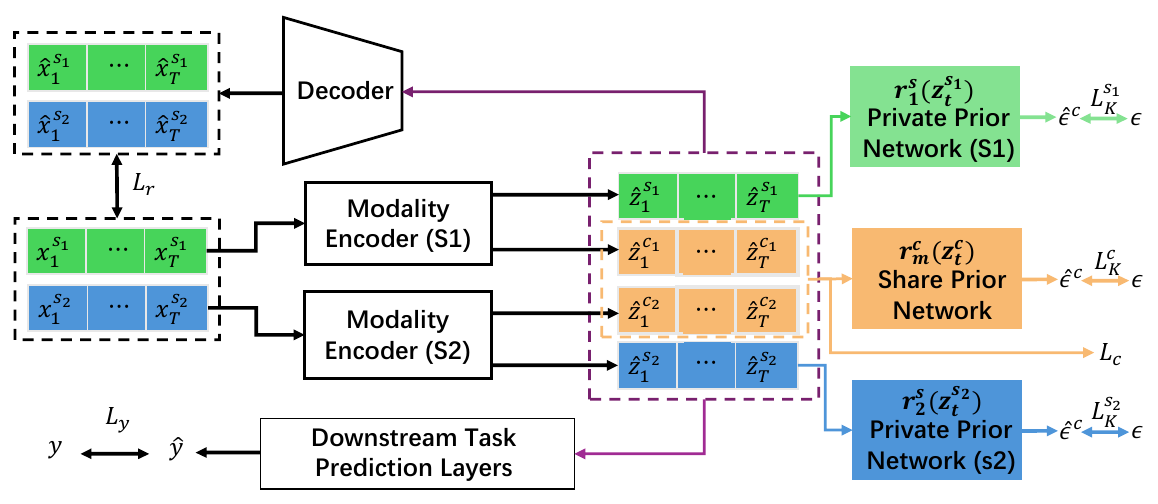}
    % \vspace{-2mm}
    \caption{Illustration of the proposed MATE model, we consider two modalities for a convenient understanding, more modalities can be easily extended. The modality-specific encoders are used to extract the latent variables of different modalities. The specific prior networks and the shared prior network are used to estimate the prior distribution for KL divergence.}
    % \vspace{-2mm}
    \label{fig:model}
\end{figure}

Based on the data generation process in Figure \ref{fig:generation}, we proposed the Multi-modal temporal Disentanglement (MATE) model as shown in Figure \ref{fig:model}, which is built upon the variation auto-encoder. Moreover, it includes the shared prior networks and the private prior networks, which are used to preserve the dependence between the modality-specific and modality-shared latent variables. Furthermore, we devise a modality-shared constraint to enforce the invariance of modality-shared latent variables from different modalities. 
% Please refer to Appendix X for the details of our model.

\subsection{Variational-Inference-based Neural Architecture}
We begin with the evidence lower bound (ELBO) based on the proposed data generation process. Without loss of generality, we consider two modalities, i.e., $M=2$, so the ELBO can be formalized as Equation (\ref{equ:elbo}). Please refer to Appendix \ref{app:elbo} for the details of derivation.
\begin{equation}
 \small
\label{equ:elbo}
\begin{split}
    p(\rvx^{s_1}_{1:T},\rvx^{s_2}_{1:T})&\geq \mathcal{L}_r- \underbrace{D_{KL}(q(\rvz^{c}_{1:T}|\rvx^{s_1}_{1:T},\rvx^{s_2}_{1:T})||p(\rvz^{c}_{1:T}))}_{\mathcal{L}_{c}} - \underbrace{D_{KL}(q(\rvz^{s_1}_{1:T}|\rvx^{s_1}_{1:T},\rvz^{c}_{1:T})||p(\rvz^{s_1}_{1:T}|\rvz^{c}_{1:T}))}_{\mathcal{L}_{s_1}}
\\&- \underbrace{D_{KL}(q(\rvz^{s_2}_{1:T}|\rvx^{s_2}_{1:T},\rvz^{c}_{1:T})||p(\rvz^{s_2}_{1:T}|\rvz^{c}_{1:T}))}_{\mathcal{L}_{s_2}},
\end{split}
\end{equation}
and $\mathcal{L}_r$ denotes the reconstruct loss and it can be formalized as:
\begin{equation}
 \small
\begin{split}
    \mathcal{L}_r =& 
\mathbb{E}_{q(\rvz^{s_1}_{1:T}|\rvx^{s_1}_{1:T},\rvz^{c}_{1:T}))}
\mathbb{E}_{q(\rvz^c_{1:T}|\rvx^{s_1}_{1:T},\rvx^{s_2}_{1:T})}
 \ln p(\rvx^{s_1}_{1:T}|\rvz^{s_1}_{1:T},\rvz^{c}_{1:T}) \\&+ 
\mathbb{E}_{q(\rvz^{s_2}_{1:T}|\rvx^{s_2}_{1:T},\rvz^{c}_{1:T})}
\mathbb{E}_{q(\rvz^c_{1:T}|\rvx^{s_1}_{1:T},\rvx^{s_2}_{1:T})}\ln p(\rvx^{s_2}_{1:T}|\rvz^{s_2}_{1:T},\rvz^{c}_{1:T})),
\end{split}
\end{equation}
where $q(\rvz_{1:T}^{s_1}|\rvx_{1:T}^{s_1},\rvz_{1:T}^{c}), q(\rvz_{1:T}^{s_2}|\rvx_{1:T}^{s_2}\rvz_{1:T}^{c})$, and $q(\rvz_{1:T}^{c}|\rvx_{1:T}^{s_1},\rvx_{1:T}^{s_2})$ are used to approximate the prior distributions of modality-specific and modality-shared latent variables and are implemented by neural architecture based on convolution neural networks (CNNs). In practice, we devise a modality-specific encoder for each modality, which can be formalized as follows:
\begin{equation}
 \small
    \rvz_{1:T}^{s_1}, \rvz_{1:T}^{c_1}=\psi_{s_1}(\rvx_{1:T}^{s_1}), \quad \rvz_{1:T}^{s_2}, \rvz_{1:T}^{c_2}=\psi_{s_2}(\rvx_{1:T}^{s_2}),
\end{equation}
Moreover, since $\rvz_{1:T}^{c_1}$ and $\rvz_{1:T}^{c_2}$ should be as similar as possible, we further devise a modality-shared constraint as shown in Equation (\ref{equ:sim}), which restricts the similarity of modality-shared latent variables between any two pairs of modalities.
\begin{equation}
 \small
\label{equ:sim}
    \mathcal{L}_s=\sum_{s_i,s_j,\in \mathcal{S}, i\neq j}\log \frac{{\rvz_{1:T}^{c_{s_i}}\cdot \rvz_{1:T}^{c_{s_j}}}}{|\rvz_{1:T}^{c_{s_i}}||\rvz_{1:T}^{c_{s_j}}|}
\end{equation}
By using the modality-shared constraint, we can simply let $\rvz_{1:T}^{c}=\rvz_{1:T}^{c_1}$ be the estimated modality-shared latent variables. 

As for $p(\rvx^{s_1}_{1:T}|\rvz^{s_1}_{1:T},\rvz^{c}_{1:T}))$ and $p(\rvx^{s_2}_{1:T}|\rvz^{s_2}_{1:T},\rvz^{c}_{1:T}))$, which model the generation process from latent variables to observations via Multi-layer Perceptron networks (MLPs) as shown in Equation (\ref{equ:gen}).
\begin{equation}
 \small
\label{equ:gen}
    \hat{\rvx}_{1:T}^{s_1}=\phi_{s_1}(\rvz_{1:T}^{s_1},\rvz_{1:T}^{c}),\quad \hat{\rvx}_{1:T}^{s_2}=\phi_{s_2}(\rvz_{1:T}^{s_2},\rvz_{1:T}^{c})
\end{equation}

Finally, the $p(\rvz_{1:T}^{s_1}|\rvz_{1:T}^c), p(\rvz_{1:T}^{s_2}|\rvz_{1:T}^c)$ and $p(\rvz_{1:T}^{c})$ in Equation (\ref{equ:elbo}) denotes the prior distribution of latent variables, which are introduced in subsection \ref{sec:prior_est}. Please refer to Appendix \ref{app:a5} for more details on the architecture of the proposed \textbf{MATE} model.
\subsection{Specific and Shared Prior Networks}\label{sec:prior_est}
\textbf{Shared Prior Networks for Modality-shared Estimation: } To model the shared prior distribution $p(\rvz_{1:T}^{c})$, we first review the transition function of shared latent variables in Equation (\ref{equ:g2}). Without loss of generality, we consider the time-lag as $1$, hence we let $\{r_i^c\}$ be a set of inverse transition functions that take $z_{t,i}^c, \rvz_{t-1}^c$ as input and output the independent noise, i.e., $\epsilon_{t,i}^c=r_i^c(z_{t,i}^c, \rvz_{t-1}^c)$. Note that these inverse transition functions can be implemented by simple MLPs. Sequentially, we devise a transformation $\sigma^c:= \{\hat{\rvz}_{t-1}^c, \hat{\rvz}_t^c\}\rightarrow\{\hat{\rvz}_{t-1}^c, \hat{\epsilon}_{t}^c\}$ and its corresponding Jacobian can be formalized as $   \mathbf{J}_{\sigma^c}=
    \begin{pmatrix}
        \mathbb{I}&0\\
        * & \text{diag}\left(\frac{\partial r^c_i}{\partial \hat{z}_{t,i}^c}\right)
    \end{pmatrix},$ 
where $*$ denotes a matrix. By applying the change of variables formula, we have the following equation, we estimated the prior distribution as follows:
\begin{equation}
 \small
\label{equ:prior_c1}
    \log p(\hat{\rvz}^c_{t-1},\hat{\rvz}^c_t)=\log p(\hat{\rvz}^c_{t-1},\hat{\epsilon}_t^c) + \log|\text{det}(\mathbf{J}_{\sigma^c})|.
\end{equation}
Moreover, we can rewrite Equation (\ref{equ:prior_c1}) to Equation (\ref{equ:prior_c2}) by using independent noise assumption.
\begin{equation}
\small
\label{equ:prior_c2}
\log p(\hat{\rvz}^c_{t}|\hat{\rvz}^c_{t-1})=\log p(\hat{\epsilon}^c_{t}) + \sum_{i=1}^{n_c}\log|\frac{\partial r_i^c}{\partial \hat{z}^c_{t,i}}|.
\end{equation}
As a result, the prior distribution shared latent variables can be estimated as follows:
\begin{equation}
 \small
\label{equ:prior_c3}
p(\hat{\rvz}_{1:T}^c)=p(\hat{\rvz}^c_1)\prod_{\tau=2}^T \left( \sum_{i=1}^{n_c}\log p(\hat{\epsilon}^c_{\tau,i}) +\sum_{i=1}^{n_c} \log|\frac{\partial r_i^c}{\partial \hat{z}^c_{\tau,i}}| \right),
\end{equation}
where $p(\hat{\epsilon}^c_{\tau,i})$ is assumed to  follow a standard Gaussian distribution. 

\textbf{Private Prior Networks for Modality-private Prior Estimation: }We assign each modality an individual prior network and take modality $s_1$ as an example. Similar to the derivation of the shared prior networks, we let $\{r_i^{s_1}\}$ be a set of inverse transition functions that take $z_{t,i}^{s_1}, \rvz_{t-1}^{s_1}$ and $\rvz_{t}^c$ as input and output the independent noise, i.e., $\epsilon_{t,i}^{s_1}=r_i^{s_1}(z_{t,i}^{s_1}, \rvz_{t-1}^{s_1},\rvz_{t}^c)$. Therefore, we can estimate the prior distribution of specific latent variables in a similar manner as shown in Equation (\ref{equ:prior_s1}).
\begin{equation}
 \small
\label{equ:prior_s1}
p(\hat{\rvz}_{1:T}^{s_1}|\hat{\rvz}^{c}_{1:T})=p(\hat{\rvz}^{s_1}_1|\hat{\rvz}^{c}_{1:T})\prod_{\tau=2}^T \left( \sum_{i=n_c+1}^n\log p(\hat{\epsilon}^{s_1}_{\tau,i}|\hat{\rvz}^{c}_{1:T}) +\sum_{i=n_c+1}^n \log|\frac{\partial r_i^{s_1}}{\partial \hat{z}^{s_1}_{\tau,i}}| \right).
\end{equation}
% By combining Equation (\ref{equ:prior_c3}), (\ref{equ:prior_s1}) and (\ref{equ:elbo}), we can model the joint distribution of multi-modal time series data.

\subsection{Model Summary}
By using the estimating private and shared priors to calculate the KL divergence in Equation (\ref{equ:elbo}), we can reconstruct the latent variables by modeling the observations from different modalities. Note that our method can be considered a flexible backbone architecture for multi-modal time series data, the learned latent variables can be applied to any downstream tasks. Therefore, by letting $\mathcal{L}_y$ be the objective function of a downstream task and combining Equation (\ref{equ:elbo}) with the modality-shared constrain in Equation (\ref{equ:sim}), the total loss of the proposed MATE model can be formalized as follows:
\begin{equation}
 \small
    \mathcal{L}_{total} = -\alpha\mathcal{L}_r + \beta (\mathcal{L}_c + \mathcal{L}_{s_1} + \mathcal{L}_{s_2}) + \gamma \mathcal{L}_s + \mathcal{L}_y,
\end{equation}
where $\alpha, \beta$ and $\gamma$ are hyper-parameters. 

% Similarly, we consider another transformation $\sigma^{s_1}:= \{\hat{\rvz}_{t-1}^c, \hat{\rvz}_t^{s_1}\}\rightarrow\{\hat{\rvz}_{t-1}^{s_1}, \hat{\epsilon}_{t}^{s_1}\}$ and its corresponding Jacobian can be formalized as follows: 

% \subsection{Model Summary}

\section{Theoretical Analysis}\label{sec:the}

To show the proposed method can learn the disentangled representation, we first provide the definition of subspace and component-wise identifiability. We further provide theoretical analysis regarding identifiability. Specifically, we leverage nonlinear ICA to show the subspace-identifiability (Theorem 1) and component-wise identifiability (Corollary 1.1) of the proposed method.

\subsection{Subspace Identifiability and Component-wise Identifiability}
Before introducing the theoretical results about identifiability, we first provide a brief introduction to subspace identification and component-wise identification. As for subspace identification \cite{li2024subspace}, the subspace identification of latent variables $\rvz_t$ means that for each ground-truth $z_{t,i}$, there exits $\hat{\rvz}_t$ and an invertible function $h_i:\mathbb{R}^n\rightarrow \mathbb{R}$, such that $z_{t,i}=h_i(\hat{\rvz}_{t})$. As for component-wise identifiability \cite{kong2022partial}, the component-wise identifiability of $\rvz_{t,i}$ means that for each ground-truth $z_{t,i}$, there exits $\hat{z}_{t,j}$ and an invertible function $h_i:\mathbb{R}\rightarrow \mathbb{R}$, such that $z_{t,i}=h_i(\hat{z}_{t,j})$. Note that the subspace  identifiability provides a coarse-grained theoretical guarantee for representation learning, ensuring that all the information is preserved. While the component-wise identifiability provides a coarse fine theoretcial guarantee, ensuring that the estimated and ground-truth latent variables are one-to-one coresponding.

\subsection{Subspace Identifiability of Latent Variables}
Based on the definition of latent causal process, we first show that the modality-shared and modality-specific latent variables are subspace identifiable, i.e., the estimated modality-shared latent variables $\hat{\rvz}_{t}^c$ (modality-specific latent variables $\hat{\rvz}_t^{s_m}$) contains all and only information in the true modality-shared latent variables ${\rvz_t^{c}}$ (modality-specific latent variables $\rvz_t^{s_m}$). Since the multi-modal time series data are pair-wise, without loss of generality, we consider modality $s_m$ as the example.
\begin{theorem}
(\textbf{Subspace Identification of the Modality-shared and Modality-specific Latent Variables}) Suppose that the observed data from different modalities is generated following the data generation process in Figure \ref{fig:generation}, and we further make the following assumptions:
\begin{itemize}[leftmargin=*]
    \item A1 \underline{(Smooth and Positive Density:)} The probability density of latent variables is smooth and positive, i.e., $p(\rvz_t|\rvz_{t-1})>0$ over $\mathcal{Z}_t$ and $\mathcal{Z}_{t-1}$.
    \item A2 \underline{(Conditional Independence:)} Conditioned on $\rvz_{t-1}$, each $z_{t,i}^c$ is independent of $z_{t,j}^c$ for $i,j \in \{1,\cdots,n_c\}, i\neq j$. And conditioned on $\rvz_{t-1}$ and $\rvz_{t}^c$, each $z_{t,i}^{s_m}$ is independent of $z_{t,j}^{s_m}$, for $i,j \in \{n_c+ 1,\cdots,n\}, i\neq j$.
    % \item A3 \underline{(Independent Noise:)}
    \item A3 \underline{(non-singular Jacobian):} Each $g_m$ has non-singular Jacobian matrices almost anywhere and $g_m$ is invertible. 
    \item A4 \underline{(Linear Independence:)} 
    For any $z_t^{s_*} \in \mathcal{Z}_{t}^{s_*}$, there exist $n_{c}+1$ values of $\rvz_{t-1,k}^{s_m}, k=n_{c}+1,\cdots,n$, such that these vectors $\bm{v}_{t,j}$ are linearly independent, where $\bm{v}_{t,j,k}$ are defined as follows:
    \begin{equation}
    \small
    \begin{split}
        \bm{v}_{t,j}=\Big(\frac{\partial^2 \log p({z}_{t,j}^{s_m}|\rvz^m_{t-1},{\rvz}_{t}^c)}{\partial z_{t,j}^{s_m}\partial z_{t-1,n_c+1}^{s_m}},\cdots,\frac{\partial^2 \log p({z}_{t,j}^{s_m}|\rvz^m_{t-1},{\rvz}_{t}^c)}{\partial z_{t,j}^{s_m}\partial z_{t-1,n}^{s_m}}\Big)
    \end{split}
    \end{equation} 
\end{itemize}
Then if $\hat{g}_1:\mathcal{Z}_t^c\times\mathcal{Z}_t^{s_1}\rightarrow \mathcal{X}_t^{s_1}$ and $\hat{g}_2:\mathcal{Z}_t^c\times\mathcal{Z}_t^{s_2}\rightarrow \mathcal{X}_t^{s_2}$ assume the generating process of the true model $(g_1,g_2)$ and match the joint distribution $p(\rvx_t^{s_1}, \rvx_t^{s_2})$ of each time step then $\rvz^c_t$ and $\rvz_t^{s_m}$ are subspace identifiable.
\end{theorem}

\textbf{Proof Sketch:} The proof can be found in Appendix \ref{app:a1}. First, we construct an invertible transformation $h_m$ between the ground-truth latent variables and estimated ones. Sequentially, we prove that the ground truth modality-shared latent variables are not the function of modality-specific latent variables by leveraging the pairing time series from different modalities. Sequentially, we leverage sufficient variability of historical information to show that the modality-specific latent variables are not the function of the estimated modality-shared latent variables. Moreover, by leveraging the invertibility of transformation $h_m$, we can obtain the Jacobian of $h_m$ as shown in Equation (\ref{equ:Jh1}),
% \begin{equation}
%  \small
% \begin{gathered}
% \label{equ:Jh1}
%     \mJ_{h_1}=\begin{bmatrix}
%     \begin{array}{c|c}
%         \textbf{A}:=\frac{\partial \rvz_t^{c}}{\partial \hat{\rvz}_t^{c}} & \textbf{B}:=\frac{\partial \rvz_t^c}{\partial \hat{\rvz}_t^{s_1}}=0 \\ \midrule
%         \textbf{C}:=\frac{\partial \rvz_t^{s_1}}{\partial \hat{\rvz}_t^c}=0 & \textbf{D}:=\frac{\partial \rvz_t^{s_1}}{\partial \hat{\rvz}_t^{s_1}},
%     \end{array}
%     \end{bmatrix}
% \end{gathered}
% \end{equation}
% \lipsum[1]
\begin{wrapfigure}{r}{0.5\textwidth} % 左侧环绕，宽度为 0.5 倍的文本宽度
\begin{equation}
 \small
\begin{gathered}
\label{equ:Jh1}
    \mJ_{h_m}=\begin{bmatrix}
    \begin{array}{c|c}
        \textbf{A}:=\frac{\partial \rvz_t^{c}}{\partial \hat{\rvz}_t^{c}} & \textbf{B}:=\frac{\partial \rvz_t^c}{\partial \hat{\rvz}_t^{s_m}}=0 \\ \midrule
        \textbf{C}:=\frac{\partial \rvz_t^{s_m}}{\partial \hat{\rvz}_t^c}=0 & \textbf{D}:=\frac{\partial \rvz_t^{s_m}}{\partial \hat{\rvz}_t^{s_m}},
    \end{array}
    \end{bmatrix}
\end{gathered}
\end{equation}
\end{wrapfigure}
% \lipsum[2]
where $B=0$ and $C=0$, since the ground truth modality-shared latent variables are not the function of modality-specific latent variables and the modality-specific latent variables are not the function of the estimated modality-shared latent variables, respectively.

\textbf{Discussion of the Assumptions:} The proof can be found in Appendix \ref{app:a1}. The first and the second assumptions are common in the existing identification results \cite{yao2022temporally,yao2021learning}. The third assumption is also common in \cite{kong2023understanding}, meaning that the influence from each latent source to observation is independence. The final assumption means that the historical information changes sufficiently, which can be easily satisfied with sufficient time series data. 

% The final assumption is similar to the existing subspace identification results like SIG \cite{li2024subspace}. Note that SIG is devised for multi-domain data with observed auxiliary variables. In the meanwhile, our method is devised for multi-modal time series data without extra auxiliary variables. 

% \textbf{Proof Sketch:}

\subsection{Component-wise Identifiability of Modality-shared Latent Variables}
Based on Theorem 1, we further establish the component-wise identifiability result as follows.
\begin{corollary}
{(\textbf{Component-wise Identification of the Modality-shared and Modality-specific Latent Variables})} Suppose that the observed data from different modalities is generated following the data generation process in Figure \ref{fig:generation}, and we further make the assumption A1, A2 and the  following assumptions:
\begin{itemize}[leftmargin=*]
    % \item A1 \underline{(Smooth and Positive Density:)} The probability density of latent variables is smooth and positive, i.e., $p(\rvz_t|\rvz_{t-1})>0$ over $\mathcal{Z}_t$ and $\mathcal{Z}_{t-1}$.
    % \item A2 \underline{(Conditional Independence:)} Conditioned on $\rvz_{t-1}$, each $z_{t,i}^c$ is independent of $z_{t,j}^c$ for $i,j \in \{1,\cdots,n_m\}, i\neq j$. And conditioned on $\rvz_{t-1}$ and $\rvz_{t}^c$, each $z_{t,i}^{s_m}$ is independent of $z_{t,j}^{s_m}$, for $i,j \in \{n_m+1,\cdots,n\}, i\neq j$.
    % \item A3 \underline{(Independent Noise:)}
    % \item A3 \underline{(non-singular Jacobian):} Each $g_m$ has non-singular Jacobian matrices almost anywhere and $g_m$ is invertible. 
    \item A5 \underline{(Linear Independence:)} 
    For any $z_t \in \mathcal{Z}_{t}$, there exist $2n +1$ values of $\rvz^{m}_{t-1,k}, k=1,\cdots,n$, such that these vectors $\bm{v}_{t,l}$ are linearly independent, where $\bm{v}_{t,l}$ are defined as follows:
    \begin{equation}
    \small
    \begin{split}
        \bm{v}_{t,l}=\Big(\frac{\partial^3 \log p({z}_{t,l}^c|\rvz^m_{t-1})}{\partial^2 z_{t,l}^c\partial z^m_{t-1,1}},\cdots,\frac{\partial^3 \log p({z}_{t,l}^c|\rvz^m_{t-1})}{\partial^2 z_{t,l}^c\partial z^m_{t-1,n}},\frac{\partial^2 \log p({z}_{t,l}^c|\rvz^m_{t-1})}{\partial z_{t,l}^c\partial z^m_{t-1,1}},\cdots,\frac{\partial^2 \log p({z}_{t,l}^c|\rvz^m_{t-1})}{\partial z_{t,l}^c\partial z^m_{t-1,n}},
     \\\frac{\partial^3 \log p({z}_{t,l}^{s_m}|\rvz^m_{t-1},{\rvz}_{t}^c)}{\partial^2 z_{t,l}^{s_m}\partial z^m_{t-1,1}},\cdots,\frac{\partial^3 \log p({z}_{t,l}^{s_m}|\rvz^m_{t-1},{\rvz}_{t}^c)}{\partial^2 z_{t,l}^{s_m}\partial z^m_{t-1,n}},\frac{\partial^2 \log p({z}_{t,l}^{s_m}|\rvz^m_{t-1},{\rvz}_{t}^c)}{\partial z_{t,l}^{s_m}\partial z^m_{t-1,1}},\cdots,\frac{\partial^2 \log p({z}_{t,l}^{s_m}|\rvz^m_{t-1},{\rvz}_{t}^c)}{\partial z_{t,l}^{s_m}\partial z^m_{t-1,n}}\Big)
    \end{split}
    \end{equation} 
\end{itemize}
Then if $\hat{g}_1:\mathcal{Z}_t^c\times\mathcal{Z}_t^{s_1}\rightarrow \mathcal{X}_t^{s_1}$ and $\hat{g}_2:\mathcal{Z}_t^c\times\mathcal{Z}_t^{s_2}\rightarrow \mathcal{X}_t^{s_2}$ assume the generating process of the true model $(g_1,g_2)$ and match the joint distribution $p(\rvx_t^{s_1}, \rvx_t^{s_2})$ of each time step then $\rvz^c_t$ is component-wise identifiable.
\end{corollary}
\textbf{Proof Sketch and Discussion:}  The proof can be found in Appendix \ref{app:a2}. Based on Theorem 1, we employ similar assumptions like \cite{yao2021learning,yao2022temporally} to construct a full-rank linear system with only zero solution, which ensures the component-wise identifiability of latent variables, i.e., the estimated and ground truth latent variables are one-to-one corresponding. 

\subsection{Relationships between Identifiability and Representation Learning} Intuitively, the proposed method is more general since existing methods with orthogonal latent space are a special case of the data generation process shown in Figure \ref{fig:generation}. We further discuss how these identifiability results benefit the representation learning for multi-modal time-series sensing signals. First, the subspace identifiability results show that the modality-shared and modality-specific latent variables are disentangled under the dependent latent process, naturally boosting the downstream tasks that require modality-shared representations. Second, the component-wise identifiability result uncovers the latent causal mechanisms of multi-modal time series data, which potentially provides the interpretability for multi-modal representation learning, i.e., finding the unobserved confounders. Third, by identifying the latent variables, we can further model the data generation process, which enhances the robustness of the representation of multi-modal time series sensing signals.

% Third, intuitively, the proposed method is more general since existing methods with orthogonal latent space is a special case of the data generation process shown in Figure \ref{fig:generation}, hence our methods can handle more real-world scenarios and extract more robust representations.

% 1. 理论保证了信息没有混淆，可以获取解耦的模态共有/特有的隐变量
% 2. 本文提出的情况更加general，正交的情况是本文的一种特例
% 3. component-wise可识别性理论有助于揭示数据背后的因果机制，有助于为揭示不同模态背后的因果关系，为提供可解析性提供了可能

% \clearpage
\begin{table*}[t]
\centering
\caption{Time series classification for Motion, Seizure, WIFI, and KETI datasets.}
\renewcommand{\arraystretch}{0.85}
\label{tab:exp_class1}
\resizebox{\textwidth}{!}{
\begin{tabular}{c|cc|cc|cc|cc}
\toprule
    & \multicolumn{2}{c|}{Motion}   
    & \multicolumn{2}{c|}{DINAMO} 
    & \multicolumn{2}{c|}{WIFI} 
    & \multicolumn{2}{c}{KETI} \\ \midrule
\textbf{Model}    & Accuracy        & Macro-F1  & Accuracy      & Macro-F1       & Accuracy            & Macro-F1
&Accuracy  &Macro-F1\\   
    \midrule
    \textbf{ResNet}        &
    89.96&	91.41&	91.88&   65.00&   90.29&	88.14&	96.05&	84.59\\
    \textbf{MaCNN}        & 
    85.57&	86.93&	90.17&	48.56&	88.81&	87.80&	93.05&	71.93
\\
    \textbf{SenenHAR}        & 
    88.95&	88.66&	89.56&	47.23&	94.63&	92.75&	96.43&	84.74
 \\
    \textbf{STFNets}        & 
    89.07&	88.84&	90.51&	47.50&	80.52&	75.93&	89.21&	69.55
 \\
    \textbf{RFNet-base}        & 
    89.93&	91.70&	90.76&	58.79&	86.31&	82.56&	95.12&	81.45
 \\
    \textbf{THAT}        & 
    89.66&	91.38&	92.76&	71.64&	95.59&	94.86&	96.33&	85.12
 \\
    \textbf{LaxCat}        &  
    60.25&	41.01&	90.64&	54.56&	76.36&	73.85&	93.33&	70.67
\\
    \textbf{UniTS}        & 
    91.02&	92.73&	90.88&	58.39&	95.83&	94.49&	96.04&	84.08
 \\
    \textbf{COCOA}        & 
    88.31&	89.27&	90.69&	55.00&	87.76&	84.51&	92.68&	74.72
\\
    \textbf{FOCAL}        &
    89.37&	90.91&	90.52&	52.00&	94.15&	92.68&	94.88&	78.47
 \\
    \textbf{CroSSL}        & 
    91.32&	89.94&	91.05&	53.13&	76.80&	68.45&	93.63&	76.25
 \\
    \midrule
    \textbf{MATE}        &  
    \textbf{92.44}&	
    \textbf{93.75}&	
    \textbf{93.31}&	
    \textbf{73.72}&
    \textbf{96.95}&	
    \textbf{96.20}&	
    \textbf{97.00}&	
    \textbf{86.93}
 \\

    \bottomrule
    \end{tabular}%
  \label{tab:exp_class1}%
}
% \vspace{-2mm}
\end{table*}%
\begin{table*}[t]
\centering
\caption{Time series classification for human motion prediction and healthcare datasets.}
\renewcommand{\arraystretch}{0.85}
\label{tab:exp_class2}
\resizebox{\textwidth}{!}{
\begin{tabular}{c|cc|cc|cc|cc}
\toprule
    & \multicolumn{2}{c|}{HumanEVA} 
    & \multicolumn{2}{c|}{H36M} 
    & \multicolumn{2}{c|}{UCIHAR} 
    & \multicolumn{2}{c}{MIT-BIH} 
 \\ \midrule
\textbf{Model}    & Accuracy        & Macro-F1  & Accuracy      & Macro-F1       & Accuracy            & Macro-F1
&Accuracy  &Macro-F1\\   
    \midrule
        \textbf{ResNet}        &
        86.68&	86.51&	92.44&	92.27&	93.12&	93.01&	98.52&	97.62	
\\
    \textbf{MaCNN}        & 
    86.27&	86.12&	78.54&	77.73&	84.57&	84.06&	97.26&	96.07	
\\
    \textbf{SenenHAR}        & 
    85.77&	86.00&	67.69&	67.44&	87.77&	87.47&	95.82&	94.79	
 \\
    \textbf{STFNets}        & 
    86.07&	85.76&	61.67&	57.20&	81.64&	81.64&	91.63&	88.97
 \\
    \textbf{RFNet-base}        & 
    97.15&	96.18&	94.14&	93.14&	95.63&	95.16&	98.64&	97.85	
 \\
    \textbf{THAT}        & 
    85.95&	85.90&	81.28&	81.27&	93.06&	93.06&	98.49&	97.56	
 \\
    \textbf{LaxCat}        &  
    86.28&	86.20&	86.09&	85.84&	89.00&	88.78&	97.77&	96.77	
\\
    \textbf{UniTS}        & 
    97.90&	97.52&	94.96&	94.81&	94.75&	94.72&	98.75&	97.95	
 \\
    \textbf{COCOA}        & 
    93.46&	91.63&	84.12&	83.85&	94.11&	93.96&	97.76&	96.64	
\\
    \textbf{FOCAL}        &
    92.15&	91.83&	89.73&	89.30&	94.36&	94.36&	98.67&	97.84	
 \\
    \textbf{CroSSL}        & 
   86.29&	86.06&	87.35&	83.62&	94.45&	93.83&	97.96&	95.06	
 \\
    \midrule
    \textbf{MATE}        &  
\textbf{98.90}&	\textbf{98.82}&	\textbf{96.12}&	\textbf{95.99}&	\textbf{95.97}&	\textbf{95.93}&	\textbf{98.97}&	\textbf{98.34}	
 \\

    \bottomrule
    \end{tabular}%
  \label{tab:exp_class2}%
}
\end{table*}%

% \section{Multi-modality Temporal Disentanglement Model}
% \vspace{-1mm}

\section{Experiments}
\subsection{Experiment Setup}\label{exp_setup}
\textbf{Datasets:} To evaluate the effectiveness of our method, we consider the different downstream tasks: classification, KNN evaluation, and linear probing on several multi-modality time series classification datasets. Specifically, we consider the WIFI \cite{yousefi2017survey}, and KETI \cite{hong2017high} datasets. Moreover, we further consider the human motion prediction datasets like Motion \cite{roggen2010collecting}, HumanEva-I \cite{sigal2010humaneva}, H36M \cite{ionescu2013human3}, UCIHAR \cite{anguita2013public}, PAMAP2 \cite{reiss2012introducing}, and RealWorld-HAR \cite{sztyler2016body}, which consider different positions of the human body as different modalities. Moreover, we also consider two healthcare datasets such as MIT-BIH \cite{moody2001impact} and D1NAMO \cite{dubosson2018open}, which are related to arrhythmia and noninvasive type 1 diabetes. Please refer to Appendix \ref{ap:a61} for more details on the dataset descriptions. 

\textbf{Evaluation Metric. } We use ADAM optimizer \cite{kingma2014adam} in all experiments and report the accuracy and the Macro-F1 as evaluation metrics. All experiments are implemented by Pytorch on a single NVIDIA RTX A100 40GB GPU. Please refer to Appendix \ref{app:a5} for the details of the model implementation.

\textbf{Baselines.} To evaluate the performance of the proposed MATE, we consider the different types of baselines. We first consider the convention ResNet \cite{he2016deep}. Sequentially, we consider several baselines for multi-modal sensing data like STFNets \cite{yao2019stfnets}, THAT \cite{li2021two}, LaxCat \cite{hsieh2021explainable}, UniTS \cite{li2021units}, and RFNet \cite{ding2020rf}. Moreover, we also consider methods based on contrastive learning like MaCNN \cite{radu2018multimodal}, 
 SenseHAR\cite{jeyakumar2019sensehar}, CPC\cite{oord2018representation}, SimCLR\cite{chen2020simple}, TS-TCC\cite{eldele2023self}, Cocoa\cite{deldari2022cocoa}, TS2Vec\cite{yue2022ts2vec}, Mixing-up\cite{wickstrom2022mixing}, TFC \cite{deldari2022cocoa}, and CroSSL \cite{deldari2024crossl}. Finally, we consider the recently proposed FOCAL \cite{liu2024focal} which considers an orthogonal latent space between domain-shared and domain-specific latent variables.
% \vspace{-9mm}
\subsection{Results and Discussion}
% \vspace{-5mm}
\textbf{Time Series Classification}: Experimental results for time series classification are shown in Table \ref{tab:exp_class1} and \ref{tab:exp_class2}. According to the experiment results, we can find that the proposed MATE model achieves the best accuracy and F1 score across different datasets. Compared with the methods based on contrastive learning and the conventional supervised learning methods, the contrastive-learning-based methods achieve better performance since they can disentangle the modality-shared and modality-specific latent variables to some extent. Moreover, since our method explicitly considers the dependence between the modality-shared and modality-specific latent variables, it outperforms the other methods like Focal and CroSSL. More interestingly, as for the experiment results of the DINAMO datasets, our method achieves a clear improvement compared with the methods with the assumption of an orthogonal latent space, which indirectly evaluates the guess mentioned in Figure \ref{fig:dependent_example}. Please refer to Appendix \ref{app:a6}  for more experiment results. 
% Please find more experiment results in the Appendix X.

\textbf{KNN Evaluation} Following the setting of \cite{liu2024focal}, we consider both the modality-shared/modality-specific latent variables and use a KNN classifier with all available labels. Experiment results are shown in Table \ref{tab:exp_knn1}. According to the experiment results, we can find that the proposed \textbf{MATE} still outperforms the other baselines like CroSSL. This is because the representation from our method preserves the dependencies of modality-shared and modality-specific latent variables, hence the representation contains richer semantic information and finally leads to better alignment results. 
\begin{wraptable}{r}{9cm}
    % \centering
\small
\caption{KNN evaluation results on Realworld-HAR and PAMAP2 datasets.}
\renewcommand{\arraystretch}{0.60}
% \label{tab:ogb_cls1}
\begin{tabular}{c|cc|cc}
\toprule
    & \multicolumn{2}{c|}{RealWorld-HAR}   
    & \multicolumn{2}{c}{PAMAP2} 
 \\ \midrule
\textbf{Model}    & Accuracy        & Macro-F1  & Accuracy      & Macro-F1      \\   
    \midrule
    \textbf{CPC}        & 
   88.94&	90.30& 89.19&87.92
 \\
     \textbf{SimCLR}        & 
   89.24&	90.64&	91.87&	91.06
 \\
     \textbf{TS-TCC}        & 
   89.47&	90.71&	92.19&	91.35
 \\
    \textbf{COCOA}        & 
   85.90&	85.79&	88.52&	87.99
 \\
    \textbf{TS2Vec}        & 
    70.25&	62.39&	56.21&	47.09
 \\
    \textbf{Mixing-up}        & 
    85.34&	86.41&	92.28&	90.95
 \\
    \textbf{TFC}        & 
    81.58&	78.73&	72.37&	63.52

\\
    \textbf{FOCAL}        &
    89.62&	90.18&	
    94.17&	93.01
 \\
    \textbf{CroSSL}        & 
   85.90&	85.69&	83.83&	83.63
 \\
    \midrule
    \textbf{MATE}        &  
    \textbf{91.66}&	
    \textbf{92.79}&	
    \textbf{94.75}&	
    \textbf{94.76}	
 \\
    \bottomrule
    \end{tabular}%
  \label{tab:exp_knn1}%
\end{wraptable}
\textbf{Linear Probing} We consider the linear probing task with four different label ratios (100\%, 10\%, 5\%, and $1\%$) as shown in Table \ref{tab:KNN_results} and Table \ref{tab:KNN_resultss}. The proposed MATE still consistently outperforms the state-of-the-art baselines in different label rates. Specifically, our method achieves 0.7\% improvement with $100\%$ lables , 16\% improvement with $10\%$ labels, 18\% improvement with $5\%$ labels, and 24\% improvement with $1\%$ labels. Note that our method still achieves an ideal performance in RealWorld-HAR dataset even with only $10\%$ ratio labels, indirectly reflecting that MATE captures sufficient semantic information with limited labels.

\begin{table}[t]
% \small
\renewcommand{\arraystretch}{0.85}
\caption{Linear probing results under different label ratios on RealWorld-HAR.}
\label{tab:KNN_results}
\resizebox{\textwidth}{!}{%
\begin{tabular}{@{}c|cc|cc|cc|cc@{}}
\toprule
\multicolumn{1}{l|}{Label Ratio} & \multicolumn{2}{c|}{100\%}      & \multicolumn{2}{c|}{10\%} & \multicolumn{2}{c|}{5\%}          & \multicolumn{2}{c}{1\%} \\ \midrule
\multicolumn{1}{l|}{}            & Accuracy       & Macro-F1       & Accuracy    & Macro-F1    & Accuracy  & Macro-F1              & Accuracy   & Macro-F1   \\ \midrule
\textbf{CPC}                        & 89.47&	90.35&	79.49&	78.85&	76.62&	72.79&	49.34&	30.84      \\
\textbf{SimCLR}                        & 89.54&	90.52&	84.21&	85.32&	79.76&	78.93&	48.35&	34.59       \\
\textbf{TS-TCC}                       & 89.70&	90.71&	82.56&	84.53&	79.16&	79.91&	53.25&	39.71       \\
\textbf{Cocoa}                        & 86.83&	86.60&	65.57&	65.24&	56.58&	56.53&	44.03&	43.50       \\
\textbf{TS2Vec}              & 
70.98&	62.92&	64.77&	56.46&	62.44&	52.59&	56.16&	46.30    \\
\textbf{Mixing-up}                    & 85.34&	86.41&	77.32&	77.92&	72.34&	71.27&	53.89&	42.99     \\
\textbf{TFC}                 & 82.58&	78.73&	72.02&	64.82&	68.13&	62.15&	63.85&	54.38      \\
\textbf{FOCAL}                        & 90.21&	90.68&	88.58&	89.68&	87.28&	87.56&	79.32&	74.78                 \\
\textbf{CroSSL}                       & 87.33&	87.42&	85.74&	85.32&	81.14&	81.32&	56.46&	47.08      \\
    \midrule
    							
\textbf{MATE}                            & \textbf{90.42} & \textbf{91.59} & \textbf{90.21}   & \textbf{91.38}   & \textbf{88.96} & 
\textbf{90.28}       & \textbf{82.63}  &
\textbf{76.29}  \\ \bottomrule
\end{tabular}%
}
% \vspace{-3mm}
\end{table}
% \vspace{-4mm}
\begin{table}[t]
% \small
\renewcommand{\arraystretch}{0.85}
\caption{Linear probing results under different label ratios on UCIHAR.}
\vspace{-2mm}
\label{tab:KNN_resultss}
\resizebox{\textwidth}{!}{%
\begin{tabular}{@{}c|cc|cc|cc|cc@{}}
\toprule
\multicolumn{1}{l|}{Label Ratio} & \multicolumn{2}{c|}{100\%}      & \multicolumn{2}{c|}{10\%} & \multicolumn{2}{c|}{5\%}          & \multicolumn{2}{c}{1\%} \\ \midrule
\multicolumn{1}{l|}{}            & Accuracy       & Macro-F1       & Accuracy    & Macro-F1    & Accuracy  & Macro-F1              & Accuracy   & Macro-F1   \\ \midrule
\textbf{CPC}                        &
72.09&	71.45&	69.71&	68.63&	61.41&	60.70&	34.57&	30.49      \\
\textbf{SimClR}                        &
86.27&	86.14&	78.94&	78.35&	68.01&	67.24&	46.46&  39.20    \\
\textbf{TS-TCC}                        &
91.11&	91.09&	85.12&	84.77&	76.29&	74.45&	61.34&  58.62      \\
\textbf{Cocoa}                        &
91.76&	91.86&	67.47&	66.79&	53.83&	53.52&	33.49&	32.86      \\
\textbf{TS2Vec}              & 
70.48&	68.37&	63.22&	61.06&	62.48&	60.49&	49.18&	42.29   \\
\textbf{Mixing-up}                    & 90.23&	90.07&	86.09&	85.71&	78.56&	77.88&	33.78&	20.31     \\
\textbf{TFC}                 & 65.53&	65.27&	53.52&	45.25&	40.91&	38.67&	45.45&	44.12      \\
\textbf{FOCAL}                        & 92.94&	92.84&	89.69&	89.46&	80.80&	79.92&	67.32&	63.13                \\
\textbf{CroSSL}                       & 92.73&	92.82&	87.91&	87.80&	77.22&	76.71&	48.59&	47.46      \\
    \midrule
    														
\textbf{MATE}                      & \textbf{93.69} 
& \textbf{93.65} 
& \textbf{90.84}   
& \textbf{90.77}   
& \textbf{81.75} 
& \textbf{80.84}       
& \textbf{68.86}  
& \textbf{63.52}  
\\ \bottomrule
\end{tabular}%
}
\end{table}

\subsection{Visualization Results} 
\vspace{-1mm}
We further provide the visualization results to evaluate that the proposed method can capture the semantic information effectively, which are shown in Figure \ref{fig:tsne}. According to the visualization results, we can find that our method can form better clusters with distinguished margins, meaning that the proposed method can well disentangle the latent variables. In the meanwhile, since the other methods assume the orthogonal latent space, they can not well extract the disentangled representation, and hence results in confusing clusters with unclear margins, for example, the entanglement among the ''Walking``, ''Walking Up``, and ''Walking Down`` in Figure \ref{fig:tsne} (b) and (e).

\subsection{Ablation Studies}
\vspace{-1mm}
To evaluate the effectiveness of each loss term, we further devise four model variants as follows. a) \textbf{MATE-p}: we remove the KL divergence terms for domain-specific latent variables. b) \textbf{MATE-s}: we remove the KL divergence terms for domain-shared latent variables. c) \textbf{MATE-r}: We remove the reconstruction loss. d) \textbf{MATE-c}: We remove the modality-shared constraint. 
Experiment results of the ablation studies on the D1NAMO and Motion datasets are shown in Figure \ref{fig:ablation}. We can draw the following conclusions 1) all the loss terms play an important role in the representation learning. 2) In the D1NAMO dataset, by removing the KL divergence terms for domain-shared and domain-specific latent variables, the model performance drops, showing that these loss terms benefit the identifiability of latent variables under dependence latent space. 3) Moreover, the drop in the performance of \textbf{MATE-r} and \textbf{MATE-c} reflects that the reconstruction loss and the modality-shared constraint conducive to preserving the semantic information. 

\section{Conclusion}
\vspace{-2mm}
We propose a representation learning framework for multi-modal time series data with theoretical guarantees, which breakthroughs the conventional orthogonal latent space assumption. Based on the data generation process for multi-modal time series data with dependent latent subspace, we devise a general disentangled representation learning framework with identifiability guarantees. Compared with the existing methods, the proposed \textbf{MATE} model can learn the disentangled time series representations even in the dependent latent subspace, hence our method is closer to the real-world scenarios. Evaluation on the time series classification, KNN evaluation, and linear probing on several multi-modal time series datasets illustrate the effectiveness of our method. Our future work would focus on the more general multi-modal time series data like audio and video data.

\clearpage
\normalem
\bibliography{main}

\begin{thebibliography}{99}
\providecommand{\natexlab}[1]{#1}
\providecommand{\url}[1]{\texttt{#1}}
\expandafter\ifx\csname urlstyle\endcsname\relax
  \providecommand{\doi}[1]{doi: #1}\else
  \providecommand{\doi}{doi: \begingroup \urlstyle{rm}\Url}\fi

\bibitem[Zhang et~al.(2024{\natexlab{a}})Zhang, Wen, Zhang, Cai, Jin, Liu,
  Zhang, Liang, Pang, Song, et~al.]{zhang2024self}
Kexin Zhang, Qingsong Wen, Chaoli Zhang, Rongyao Cai, Ming Jin, Yong Liu,
  James~Y Zhang, Yuxuan Liang, Guansong Pang, Dongjin Song, et~al.
\newblock Self-supervised learning for time series analysis: Taxonomy,
  progress, and prospects.
\newblock \emph{IEEE Transactions on Pattern Analysis and Machine
  Intelligence}, 2024{\natexlab{a}}.

\bibitem[Liang et~al.(2024)Liang, Wen, Nie, Jiang, Jin, Song, Pan, and
  Wen]{liang2024foundation}
Yuxuan Liang, Haomin Wen, Yuqi Nie, Yushan Jiang, Ming Jin, Dongjin Song,
  Shirui Pan, and Qingsong Wen.
\newblock Foundation models for time series analysis: A tutorial and survey.
\newblock \emph{arXiv preprint arXiv:2403.14735}, 2024.

\bibitem[Li et~al.(2023{\natexlab{a}})Li, Cui, Zhang, Zhu, Wang, Tsang, and
  Zhou]{li2023difformer}
Bing Li, Wei Cui, Le~Zhang, Ce~Zhu, Wei Wang, Ivor Tsang, and Joey~Tianyi Zhou.
\newblock Difformer: Multi-resolutional differencing transformer with dynamic
  ranging for time series analysis.
\newblock \emph{IEEE Transactions on Pattern Analysis and Machine
  Intelligence}, 2023{\natexlab{a}}.

\bibitem[Wu et~al.(2022)Wu, Hu, Liu, Zhou, Wang, and Long]{wu2022timesnet}
Haixu Wu, Tengge Hu, Yong Liu, Hang Zhou, Jianmin Wang, and Mingsheng Long.
\newblock Timesnet: Temporal 2d-variation modeling for general time series
  analysis.
\newblock In \emph{The eleventh international conference on learning
  representations}, 2022.

\bibitem[Luo and Wang(2024)]{luo2024moderntcn}
Donghao Luo and Xue Wang.
\newblock Moderntcn: A modern pure convolution structure for general time
  series analysis.
\newblock In \emph{The Twelfth International Conference on Learning
  Representations}, 2024.

\bibitem[Zhou et~al.(2024)Zhou, Niu, Sun, Jin, et~al.]{zhou2024one}
Tian Zhou, Peisong Niu, Liang Sun, Rong Jin, et~al.
\newblock One fits all: Power general time series analysis by pretrained lm.
\newblock \emph{Advances in neural information processing systems}, 36, 2024.

\bibitem[Shao et~al.(2023)Shao, Wang, Xu, Wei, Yu, Zhang, Yao, Jin, Cao, Cong,
  et~al.]{shao2023exploring}
Zezhi Shao, Fei Wang, Yongjun Xu, Wei Wei, Chengqing Yu, Zhao Zhang, Di~Yao,
  Guangyin Jin, Xin Cao, Gao Cong, et~al.
\newblock Exploring progress in multivariate time series forecasting:
  Comprehensive benchmarking and heterogeneity analysis.
\newblock \emph{arXiv preprint arXiv:2310.06119}, 2023.

\bibitem[Liu et~al.(2024{\natexlab{a}})Liu, Zhong, Jin, Tao, Luo, Liu, and
  Zhao]{liu2024mtsa}
Chengzhi Liu, Chong Zhong, Mingyu Jin, Zheng Tao, Zihong Luo, Chenghao Liu, and
  Shuliang Zhao.
\newblock Mtsa-snn: A multi-modal time series analysis model based on spiking
  neural network.
\newblock \emph{arXiv preprint arXiv:2402.05423}, 2024{\natexlab{a}}.

\bibitem[Liu et~al.(2024{\natexlab{b}})Liu, Kimura, Liu, Wang, Li, Diggavi,
  Srivastava, and Abdelzaher]{liu2024focal}
Shengzhong Liu, Tomoyoshi Kimura, Dongxin Liu, Ruijie Wang, Jinyang Li, Suhas
  Diggavi, Mani Srivastava, and Tarek Abdelzaher.
\newblock Focal: Contrastive learning for multimodal time-series sensing
  signals in factorized orthogonal latent space.
\newblock \emph{Advances in Neural Information Processing Systems}, 36,
  2024{\natexlab{b}}.

\bibitem[Radu et~al.(2016)Radu, Lane, Bhattacharya, Mascolo, Marina, and
  Kawsar]{radu2016towards}
Valentin Radu, Nicholas~D Lane, Sourav Bhattacharya, Cecilia Mascolo, Mahesh~K
  Marina, and Fahim Kawsar.
\newblock Towards multimodal deep learning for activity recognition on mobile
  devices.
\newblock In \emph{Proceedings of the 2016 ACM International Joint Conference
  on Pervasive and Ubiquitous Computing: Adjunct}, pages 185--188, 2016.

\bibitem[Xu et~al.(2016)Xu, Yang, Zhou, Shangguan, Yi, and Liu]{xu2016indoor}
Han Xu, Zheng Yang, Zimu Zhou, Longfei Shangguan, Ke~Yi, and Yunhao Liu.
\newblock Indoor localization via multi-modal sensing on smartphones.
\newblock In \emph{Proceedings of the 2016 ACM International Joint Conference
  on Pervasive and Ubiquitous Computing}, pages 208--219, 2016.

\bibitem[Nonnenmacher et~al.(2022)Nonnenmacher, Oldenburg, Steinwart, and
  Reeb]{nonnenmacher2022utilizing}
Manuel~T Nonnenmacher, Lukas Oldenburg, Ingo Steinwart, and David Reeb.
\newblock Utilizing expert features for contrastive learning of time-series
  representations.
\newblock In \emph{International Conference on Machine Learning}, pages
  16969--16989. PMLR, 2022.

\bibitem[Zhang et~al.(2022)Zhang, Chu, Ma, Zhu, Wang, Wang, and
  Zhao]{zhang2022m3care}
Chaohe Zhang, Xu~Chu, Liantao Ma, Yinghao Zhu, Yasha Wang, Jiangtao Wang, and
  Junfeng Zhao.
\newblock M3care: Learning with missing modalities in multimodal healthcare
  data.
\newblock In \emph{Proceedings of the 28th ACM SIGKDD Conference on Knowledge
  Discovery and Data Mining}, pages 2418--2428, 2022.

\bibitem[Makarious et~al.(2022)Makarious, Leonard, Vitale, Iwaki, Sargent,
  Dadu, Violich, Hutchins, Saffo, Bandres-Ciga, et~al.]{makarious2022multi}
Mary~B Makarious, Hampton~L Leonard, Dan Vitale, Hirotaka Iwaki, Lana Sargent,
  Anant Dadu, Ivo Violich, Elizabeth Hutchins, David Saffo, Sara Bandres-Ciga,
  et~al.
\newblock Multi-modality machine learning predicting parkinson’s disease.
\newblock \emph{npj Parkinson's Disease}, 8\penalty0 (1):\penalty0 35, 2022.

\bibitem[Iglesias(2023)]{iglesias2023ready}
Juan~Eugenio Iglesias.
\newblock A ready-to-use machine learning tool for symmetric multi-modality
  registration of brain mri.
\newblock \emph{Scientific Reports}, 13\penalty0 (1):\penalty0 6657, 2023.

\bibitem[Cheng et~al.(2022)Cheng, Yang, Xiang, and Liu]{cheng2022financial}
Dawei Cheng, Fangzhou Yang, Sheng Xiang, and Jin Liu.
\newblock Financial time series forecasting with multi-modality graph neural
  network.
\newblock \emph{Pattern Recognition}, 121:\penalty0 108218, 2022.

\bibitem[Zhou et~al.(2020)Zhou, Zheng, Zhu, Li, and He]{zhou2020domain}
Dawei Zhou, Lecheng Zheng, Yada Zhu, Jianbo Li, and Jingrui He.
\newblock Domain adaptive multi-modality neural attention network for financial
  forecasting.
\newblock In \emph{Proceedings of The Web Conference 2020}, pages 2230--2240,
  2020.

\bibitem[Deldari et~al.(2022)Deldari, Xue, Saeed, Smith, and
  Salim]{deldari2022cocoa}
Shohreh Deldari, Hao Xue, Aaqib Saeed, Daniel~V Smith, and Flora~D Salim.
\newblock Cocoa: Cross modality contrastive learning for sensor data.
\newblock \emph{Proceedings of the ACM on Interactive, Mobile, Wearable and
  Ubiquitous Technologies}, 6\penalty0 (3):\penalty0 1--28, 2022.

\bibitem[Ouyang et~al.(2022)Ouyang, Shuai, Zhou, Shi, Xie, Xing, and
  Huang]{ouyang2022cosmo}
Xiaomin Ouyang, Xian Shuai, Jiayu Zhou, Ivy~Wang Shi, Zhiyuan Xie, Guoliang
  Xing, and Jianwei Huang.
\newblock Cosmo: contrastive fusion learning with small data for multimodal
  human activity recognition.
\newblock In \emph{Proceedings of the 28th Annual International Conference on
  Mobile Computing And Networking}, pages 324--337, 2022.

\bibitem[Huang et~al.(2023)Huang, Wang, Zhao, and Zheng]{huang2023latent}
Zenan Huang, Haobo Wang, Junbo Zhao, and Nenggan Zheng.
\newblock Latent processes identification from multi-view time series.
\newblock \emph{arXiv preprint arXiv:2305.08164}, 2023.

\bibitem[Li et~al.(2024)Li, Cai, Chen, Sun, Hao, and Zhang]{li2024subspace}
Zijian Li, Ruichu Cai, Guangyi Chen, Boyang Sun, Zhifeng Hao, and Kun Zhang.
\newblock Subspace identification for multi-source domain adaptation.
\newblock \emph{Advances in Neural Information Processing Systems}, 36, 2024.

\bibitem[Kong et~al.(2022)Kong, Xie, Yao, Zheng, Chen, Stojanov, Akinwande, and
  Zhang]{kong2022partial}
Lingjing Kong, Shaoan Xie, Weiran Yao, Yujia Zheng, Guangyi Chen, Petar
  Stojanov, Victor Akinwande, and Kun Zhang.
\newblock Partial disentanglement for domain adaptation.
\newblock In \emph{International conference on machine learning}, pages
  11455--11472. PMLR, 2022.

\bibitem[Yao et~al.(2022)Yao, Chen, and Zhang]{yao2022temporally}
Weiran Yao, Guangyi Chen, and Kun Zhang.
\newblock Temporally disentangled representation learning.
\newblock \emph{Advances in Neural Information Processing Systems},
  35:\penalty0 26492--26503, 2022.

\bibitem[Yao et~al.(2021)Yao, Sun, Ho, Sun, and Zhang]{yao2021learning}
Weiran Yao, Yuewen Sun, Alex Ho, Changyin Sun, and Kun Zhang.
\newblock Learning temporally causal latent processes from general temporal
  data.
\newblock \emph{arXiv preprint arXiv:2110.05428}, 2021.

\bibitem[Kong et~al.(2023{\natexlab{a}})Kong, Ma, Chen, Xing, Chi, Morency, and
  Zhang]{kong2023understanding}
Lingjing Kong, Martin~Q Ma, Guangyi Chen, Eric~P Xing, Yuejie Chi,
  Louis-Philippe Morency, and Kun Zhang.
\newblock Understanding masked autoencoders via hierarchical latent variable
  models.
\newblock In \emph{Proceedings of the IEEE/CVF Conference on Computer Vision
  and Pattern Recognition}, pages 7918--7928, 2023{\natexlab{a}}.

\bibitem[Yousefi et~al.(2017)Yousefi, Narui, Dayal, Ermon, and
  Valaee]{yousefi2017survey}
Siamak Yousefi, Hirokazu Narui, Sankalp Dayal, Stefano Ermon, and Shahrokh
  Valaee.
\newblock A survey on behavior recognition using wifi channel state
  information.
\newblock \emph{IEEE Communications Magazine}, 55\penalty0 (10):\penalty0
  98--104, 2017.

\bibitem[Hong et~al.(2017)Hong, Gu, and Whitehouse]{hong2017high}
Dezhi Hong, Quanquan Gu, and Kamin Whitehouse.
\newblock High-dimensional time series clustering via cross-predictability.
\newblock In \emph{Artificial Intelligence and Statistics}, pages 642--651.
  PMLR, 2017.

\bibitem[Roggen et~al.(2010)Roggen, Calatroni, Rossi, Holleczek, F{\"o}rster,
  Tr{\"o}ster, Lukowicz, Bannach, Pirkl, Ferscha, et~al.]{roggen2010collecting}
Daniel Roggen, Alberto Calatroni, Mirco Rossi, Thomas Holleczek, Kilian
  F{\"o}rster, Gerhard Tr{\"o}ster, Paul Lukowicz, David Bannach, Gerald Pirkl,
  Alois Ferscha, et~al.
\newblock Collecting complex activity datasets in highly rich networked sensor
  environments.
\newblock In \emph{2010 Seventh international conference on networked sensing
  systems (INSS)}, pages 233--240. IEEE, 2010.

\bibitem[Sigal et~al.(2010)Sigal, Balan, and Black]{sigal2010humaneva}
Leonid Sigal, Alexandru~O Balan, and Michael~J Black.
\newblock Humaneva: Synchronized video and motion capture dataset and baseline
  algorithm for evaluation of articulated human motion.
\newblock \emph{International journal of computer vision}, 87\penalty0
  (1):\penalty0 4--27, 2010.

\bibitem[Ionescu et~al.(2013)Ionescu, Papava, Olaru, and
  Sminchisescu]{ionescu2013human3}
Catalin Ionescu, Dragos Papava, Vlad Olaru, and Cristian Sminchisescu.
\newblock Human3. 6m: Large scale datasets and predictive methods for 3d human
  sensing in natural environments.
\newblock \emph{IEEE transactions on pattern analysis and machine
  intelligence}, 36\penalty0 (7):\penalty0 1325--1339, 2013.

\bibitem[Anguita et~al.(2013)Anguita, Ghio, Oneto, Parra, Reyes-Ortiz,
  et~al.]{anguita2013public}
Davide Anguita, Alessandro Ghio, Luca Oneto, Xavier Parra, Jorge~Luis
  Reyes-Ortiz, et~al.
\newblock A public domain dataset for human activity recognition using
  smartphones.
\newblock In \emph{Esann}, volume~3, page~3, 2013.

\bibitem[Reiss and Stricker(2012)]{reiss2012introducing}
Attila Reiss and Didier Stricker.
\newblock Introducing a new benchmarked dataset for activity monitoring.
\newblock In \emph{2012 16th international symposium on wearable computers},
  pages 108--109. IEEE, 2012.

\bibitem[Sztyler and Stuckenschmidt(2016)]{sztyler2016body}
Timo Sztyler and Heiner Stuckenschmidt.
\newblock On-body localization of wearable devices: An investigation of
  position-aware activity recognition.
\newblock In \emph{2016 IEEE International Conference on Pervasive Computing
  and Communications (PerCom)}, pages 1--9. IEEE, 2016.

\bibitem[Moody and Mark(2001)]{moody2001impact}
George~B Moody and Roger~G Mark.
\newblock The impact of the mit-bih arrhythmia database.
\newblock \emph{IEEE engineering in medicine and biology magazine}, 20\penalty0
  (3):\penalty0 45--50, 2001.

\bibitem[Dubosson et~al.(2018)Dubosson, Ranvier, Bromuri, Calbimonte, Ruiz, and
  Schumacher]{dubosson2018open}
Fabien Dubosson, Jean-Eudes Ranvier, Stefano Bromuri, Jean-Paul Calbimonte,
  Juan Ruiz, and Michael Schumacher.
\newblock The open d1namo dataset: A multi-modal dataset for research on
  non-invasive type 1 diabetes management.
\newblock \emph{Informatics in Medicine Unlocked}, 13:\penalty0 92--100, 2018.

\bibitem[Kingma and Ba(2014)]{kingma2014adam}
Diederik~P Kingma and Jimmy Ba.
\newblock Adam: A method for stochastic optimization.
\newblock \emph{arXiv preprint arXiv:1412.6980}, 2014.

\bibitem[He et~al.(2016)He, Zhang, Ren, and Sun]{he2016deep}
Kaiming He, Xiangyu Zhang, Shaoqing Ren, and Jian Sun.
\newblock Deep residual learning for image recognition.
\newblock In \emph{Proceedings of the IEEE conference on computer vision and
  pattern recognition}, pages 770--778, 2016.

\bibitem[Yao et~al.(2019)Yao, Piao, Jiang, Zhao, Shao, Liu, Liu, Li, Wang, Hu,
  et~al.]{yao2019stfnets}
Shuochao Yao, Ailing Piao, Wenjun Jiang, Yiran Zhao, Huajie Shao, Shengzhong
  Liu, Dongxin Liu, Jinyang Li, Tianshi Wang, Shaohan Hu, et~al.
\newblock Stfnets: Learning sensing signals from the time-frequency perspective
  with short-time fourier neural networks.
\newblock In \emph{The World Wide Web Conference}, pages 2192--2202, 2019.

\bibitem[Li et~al.(2021{\natexlab{a}})Li, Cui, Wang, Zhang, Chen, and
  Wu]{li2021two}
Bing Li, Wei Cui, Wei Wang, Le~Zhang, Zhenghua Chen, and Min Wu.
\newblock Two-stream convolution augmented transformer for human activity
  recognition.
\newblock In \emph{Proceedings of the AAAI Conference on Artificial
  Intelligence}, volume~35, pages 286--293, 2021{\natexlab{a}}.

\bibitem[Hsieh et~al.(2021)Hsieh, Wang, Sun, and Honavar]{hsieh2021explainable}
Tsung-Yu Hsieh, Suhang Wang, Yiwei Sun, and Vasant Honavar.
\newblock Explainable multivariate time series classification: a deep neural
  network which learns to attend to important variables as well as time
  intervals.
\newblock In \emph{Proceedings of the 14th ACM international conference on web
  search and data mining}, pages 607--615, 2021.

\bibitem[Li et~al.(2021{\natexlab{b}})Li, Chowdhury, Shang, Gupta, and
  Hong]{li2021units}
Shuheng Li, Ranak~Roy Chowdhury, Jingbo Shang, Rajesh~K Gupta, and Dezhi Hong.
\newblock Units: Short-time fourier inspired neural networks for sensory time
  series classification.
\newblock In \emph{Proceedings of the 19th ACM Conference on Embedded Networked
  Sensor Systems}, pages 234--247, 2021{\natexlab{b}}.

\bibitem[Ding et~al.(2020)Ding, Chen, Zheng, and Luo]{ding2020rf}
Shuya Ding, Zhe Chen, Tianyue Zheng, and Jun Luo.
\newblock Rf-net: A unified meta-learning framework for rf-enabled one-shot
  human activity recognition.
\newblock In \emph{Proceedings of the 18th Conference on Embedded Networked
  Sensor Systems}, pages 517--530, 2020.

\bibitem[Radu et~al.(2018)Radu, Tong, Bhattacharya, Lane, Mascolo, Marina, and
  Kawsar]{radu2018multimodal}
Valentin Radu, Catherine Tong, Sourav Bhattacharya, Nicholas~D Lane, Cecilia
  Mascolo, Mahesh~K Marina, and Fahim Kawsar.
\newblock Multimodal deep learning for activity and context recognition.
\newblock \emph{Proceedings of the ACM on interactive, mobile, wearable and
  ubiquitous technologies}, 1\penalty0 (4):\penalty0 1--27, 2018.

\bibitem[Jeyakumar et~al.(2019)Jeyakumar, Lai, Suda, and
  Srivastava]{jeyakumar2019sensehar}
Jeya~Vikranth Jeyakumar, Liangzhen Lai, Naveen Suda, and Mani Srivastava.
\newblock Sensehar: a robust virtual activity sensor for smartphones and
  wearables.
\newblock In \emph{Proceedings of the 17th Conference on Embedded Networked
  Sensor Systems}, pages 15--28, 2019.

\bibitem[Oord et~al.(2018)Oord, Li, and Vinyals]{oord2018representation}
Aaron van~den Oord, Yazhe Li, and Oriol Vinyals.
\newblock Representation learning with contrastive predictive coding.
\newblock \emph{arXiv preprint arXiv:1807.03748}, 2018.

\bibitem[Chen et~al.(2020)Chen, Kornblith, Norouzi, and Hinton]{chen2020simple}
Ting Chen, Simon Kornblith, Mohammad Norouzi, and Geoffrey Hinton.
\newblock A simple framework for contrastive learning of visual
  representations.
\newblock In \emph{International conference on machine learning}, pages
  1597--1607. PMLR, 2020.

\bibitem[Eldele et~al.(2023)Eldele, Ragab, Chen, Wu, Kwoh, Li, and
  Guan]{eldele2023self}
Emadeldeen Eldele, Mohamed Ragab, Zhenghua Chen, Min Wu, Chee-Keong Kwoh,
  Xiaoli Li, and Cuntai Guan.
\newblock Self-supervised contrastive representation learning for
  semi-supervised time-series classification.
\newblock \emph{IEEE Transactions on Pattern Analysis and Machine
  Intelligence}, 2023.

\bibitem[Yue et~al.(2022)Yue, Wang, Duan, Yang, Huang, Tong, and
  Xu]{yue2022ts2vec}
Zhihan Yue, Yujing Wang, Juanyong Duan, Tianmeng Yang, Congrui Huang, Yunhai
  Tong, and Bixiong Xu.
\newblock Ts2vec: Towards universal representation of time series.
\newblock In \emph{Proceedings of the AAAI Conference on Artificial
  Intelligence}, volume~36, pages 8980--8987, 2022.

\bibitem[Wickstr{\o}m et~al.(2022)Wickstr{\o}m, Kampffmeyer, Mikalsen, and
  Jenssen]{wickstrom2022mixing}
Kristoffer Wickstr{\o}m, Michael Kampffmeyer, Karl~{\O}yvind Mikalsen, and
  Robert Jenssen.
\newblock Mixing up contrastive learning: Self-supervised representation
  learning for time series.
\newblock \emph{Pattern Recognition Letters}, 155:\penalty0 54--61, 2022.

\bibitem[Deldari et~al.(2024)Deldari, Spathis, Malekzadeh, Kawsar, Salim, and
  Mathur]{deldari2024crossl}
Shohreh Deldari, Dimitris Spathis, Mohammad Malekzadeh, Fahim Kawsar, Flora~D
  Salim, and Akhil Mathur.
\newblock Crossl: Cross-modal self-supervised learning for time-series through
  latent masking.
\newblock In \emph{Proceedings of the 17th ACM International Conference on Web
  Search and Data Mining}, pages 152--160, 2024.

\bibitem[Jiang et~al.(2023)Jiang, Chen, Zhao, Chen, Ping, Tran, Xu, Zeng, and
  Chilimbi]{jiang2023understanding}
Qian Jiang, Changyou Chen, Han Zhao, Liqun Chen, Qing Ping, Son~Dinh Tran,
  Yi~Xu, Belinda Zeng, and Trishul Chilimbi.
\newblock Understanding and constructing latent modality structures in
  multi-modal representation learning.
\newblock In \emph{Proceedings of the IEEE/CVF Conference on Computer Vision
  and Pattern Recognition}, pages 7661--7671, 2023.

\bibitem[Liang et~al.(2022)Liang, Lyu, Fan, Tsaw, Liu, Mo, Yogatama, Morency,
  and Salakhutdinov]{liang2022high}
Paul~Pu Liang, Yiwei Lyu, Xiang Fan, Jeffrey Tsaw, Yudong Liu, Shentong Mo,
  Dani Yogatama, Louis-Philippe Morency, and Russ Salakhutdinov.
\newblock High-modality multimodal transformer: Quantifying modality \&
  interaction heterogeneity for high-modality representation learning.
\newblock \emph{Transactions on Machine Learning Research}, 2022.

\bibitem[Tu et~al.(2022)Tu, Cao, Mostafavi, Gao, et~al.]{tu2022cross}
Xinming Tu, Zhi-Jie Cao, Sara Mostafavi, Ge~Gao, et~al.
\newblock Cross-linked unified embedding for cross-modality representation
  learning.
\newblock \emph{Advances in Neural Information Processing Systems},
  35:\penalty0 15942--15955, 2022.

\bibitem[Liu et~al.(2023{\natexlab{a}})Liu, Wei, Lu, Sun, Wang, and
  Zheng]{liu2023m3ae}
Hong Liu, Dong Wei, Donghuan Lu, Jinghan Sun, Liansheng Wang, and Yefeng Zheng.
\newblock M3ae: Multimodal representation learning for brain tumor segmentation
  with missing modalities.
\newblock In \emph{Proceedings of the AAAI Conference on Artificial
  Intelligence}, volume~37, pages 1657--1665, 2023{\natexlab{a}}.

\bibitem[Radford et~al.(2021)Radford, Kim, Hallacy, Ramesh, Goh, Agarwal,
  Sastry, Askell, Mishkin, Clark, et~al.]{radford2021learning}
Alec Radford, Jong~Wook Kim, Chris Hallacy, Aditya Ramesh, Gabriel Goh,
  Sandhini Agarwal, Girish Sastry, Amanda Askell, Pamela Mishkin, Jack Clark,
  et~al.
\newblock Learning transferable visual models from natural language
  supervision.
\newblock In \emph{International conference on machine learning}, pages
  8748--8763. PMLR, 2021.

\bibitem[Duan et~al.(2022)Duan, Chen, Tran, Yang, Xu, Zeng, and
  Chilimbi]{duan2022multi}
Jiali Duan, Liqun Chen, Son Tran, Jinyu Yang, Yi~Xu, Belinda Zeng, and Trishul
  Chilimbi.
\newblock Multi-modal alignment using representation codebook.
\newblock In \emph{Proceedings of the IEEE/CVF Conference on Computer Vision
  and Pattern Recognition}, pages 15651--15660, 2022.

\bibitem[Kwon et~al.(2022)Kwon, Cai, Ravichandran, Bas, Bhotika, and
  Soatto]{kwon2022masked}
Gukyeong Kwon, Zhaowei Cai, Avinash Ravichandran, Erhan Bas, Rahul Bhotika, and
  Stefano Soatto.
\newblock Masked vision and language modeling for multi-modal representation
  learning.
\newblock \emph{arXiv preprint arXiv:2208.02131}, 2022.

\bibitem[Li et~al.(2021{\natexlab{c}})Li, Selvaraju, Gotmare, Joty, Xiong, and
  Hoi]{li2021align}
Junnan Li, Ramprasaath Selvaraju, Akhilesh Gotmare, Shafiq Joty, Caiming Xiong,
  and Steven Chu~Hong Hoi.
\newblock Align before fuse: Vision and language representation learning with
  momentum distillation.
\newblock \emph{Advances in neural information processing systems},
  34:\penalty0 9694--9705, 2021{\natexlab{c}}.

\bibitem[Shukor et~al.(2022)Shukor, Couairon, and Cord]{shukor2022efficient}
Mustafa Shukor, Guillaume Couairon, and Matthieu Cord.
\newblock Efficient vision-language pretraining with visual concepts and
  hierarchical alignment.
\newblock \emph{arXiv preprint arXiv:2208.13628}, 2022.

\bibitem[Yang et~al.(2022)Yang, Duan, Tran, Xu, Chanda, Chen, Zeng, Chilimbi,
  and Huang]{yang2022vision}
Jinyu Yang, Jiali Duan, Son Tran, Yi~Xu, Sampath Chanda, Liqun Chen, Belinda
  Zeng, Trishul Chilimbi, and Junzhou Huang.
\newblock Vision-language pre-training with triple contrastive learning.
\newblock In \emph{Proceedings of the IEEE/CVF Conference on Computer Vision
  and Pattern Recognition}, pages 15671--15680, 2022.

\bibitem[Jaiswal et~al.(2020)Jaiswal, Babu, Zadeh, Banerjee, and
  Makedon]{jaiswal2020survey}
Ashish Jaiswal, Ashwin~Ramesh Babu, Mohammad~Zaki Zadeh, Debapriya Banerjee,
  and Fillia Makedon.
\newblock A survey on contrastive self-supervised learning.
\newblock \emph{Technologies}, 9\penalty0 (1):\penalty0 2, 2020.

\bibitem[Zhai et~al.(2019)Zhai, Oliver, Kolesnikov, and Beyer]{zhai2019s4l}
Xiaohua Zhai, Avital Oliver, Alexander Kolesnikov, and Lucas Beyer.
\newblock S4l: Self-supervised semi-supervised learning.
\newblock In \emph{Proceedings of the IEEE/CVF international conference on
  computer vision}, pages 1476--1485, 2019.

\bibitem[Liu et~al.(2021)Liu, Zhang, Hou, Mian, Wang, Zhang, and
  Tang]{liu2021self}
Xiao Liu, Fanjin Zhang, Zhenyu Hou, Li~Mian, Zhaoyu Wang, Jing Zhang, and Jie
  Tang.
\newblock Self-supervised learning: Generative or contrastive.
\newblock \emph{IEEE transactions on knowledge and data engineering},
  35\penalty0 (1):\penalty0 857--876, 2021.

\bibitem[He et~al.(2022)He, Chen, Xie, Li, Doll{\'a}r, and
  Girshick]{he2022masked}
Kaiming He, Xinlei Chen, Saining Xie, Yanghao Li, Piotr Doll{\'a}r, and Ross
  Girshick.
\newblock Masked autoencoders are scalable vision learners.
\newblock In \emph{Proceedings of the IEEE/CVF conference on computer vision
  and pattern recognition}, pages 16000--16009, 2022.

\bibitem[Geng et~al.(2022)Geng, Liu, Lee, Schuurmans, Levine, and
  Abbeel]{geng2022multimodal}
Xinyang Geng, Hao Liu, Lisa Lee, Dale Schuurmans, Sergey Levine, and Pieter
  Abbeel.
\newblock Multimodal masked autoencoders learn transferable representations.
\newblock \emph{arXiv preprint arXiv:2205.14204}, 2022.

\bibitem[Limoyo et~al.(2022)Limoyo, Ablett, and Kelly]{limoyo2022learning}
Oliver Limoyo, Trevor Ablett, and Jonathan Kelly.
\newblock Learning sequential latent variable models from multimodal time
  series data.
\newblock In \emph{International Conference on Intelligent Autonomous Systems},
  pages 511--528. Springer, 2022.

\bibitem[Khattar et~al.(2019)Khattar, Goud, Gupta, and Varma]{khattar2019mvae}
Dhruv Khattar, Jaipal~Singh Goud, Manish Gupta, and Vasudeva Varma.
\newblock Mvae: Multimodal variational autoencoder for fake news detection.
\newblock In \emph{The world wide web conference}, pages 2915--2921, 2019.

\bibitem[Deng et~al.(2024)Deng, Jiang, Zhang, and Song]{deng2024multi}
Jiewen Deng, Renhe Jiang, Jiaqi Zhang, and Xuan Song.
\newblock Multi-modality spatio-temporal forecasting via self-supervised
  learning.
\newblock \emph{arXiv preprint arXiv:2405.03255}, 2024.

\bibitem[Kara et~al.(2024)Kara, Kimura, Liu, Li, Liu, Wang, Wang, Chen, Hu, and
  Abdelzaher]{kara2024freqmae}
Denizhan Kara, Tomoyoshi Kimura, Shengzhong Liu, Jinyang Li, Dongxin Liu,
  Tianshi Wang, Ruijie Wang, Yizhuo Chen, Yigong Hu, and Tarek Abdelzaher.
\newblock Freqmae: Frequency-aware masked autoencoder for multi-modal iot
  sensing.
\newblock In \emph{Proceedings of the ACM on Web Conference 2024}, pages
  2795--2806, 2024.

\bibitem[Rajendran et~al.(2024)Rajendran, Buchholz, Aragam, Sch{\"o}lkopf, and
  Ravikumar]{rajendran2024learning}
Goutham Rajendran, Simon Buchholz, Bryon Aragam, Bernhard Sch{\"o}lkopf, and
  Pradeep Ravikumar.
\newblock Learning interpretable concepts: Unifying causal representation
  learning and foundation models.
\newblock \emph{arXiv preprint arXiv:2402.09236}, 2024.

\bibitem[Mansouri et~al.(2023)Mansouri, Hartford, Zhang, and
  Bengio]{mansouri2023object}
Amin Mansouri, Jason Hartford, Yan Zhang, and Yoshua Bengio.
\newblock Object-centric architectures enable efficient causal representation
  learning.
\newblock \emph{arXiv preprint arXiv:2310.19054}, 2023.

\bibitem[Wendong et~al.(2024)Wendong, Keki{\'c}, von K{\"u}gelgen, Buchholz,
  Besserve, Gresele, and Sch{\"o}lkopf]{wendong2024causal}
Liang Wendong, Armin Keki{\'c}, Julius von K{\"u}gelgen, Simon Buchholz, Michel
  Besserve, Luigi Gresele, and Bernhard Sch{\"o}lkopf.
\newblock Causal component analysis.
\newblock \emph{Advances in Neural Information Processing Systems}, 36, 2024.

\bibitem[Yao et~al.(2023)Yao, Xu, Lachapelle, Magliacane, Taslakian, Martius,
  von K{\"u}gelgen, and Locatello]{yao2023multi}
Dingling Yao, Danru Xu, S{\'e}bastien Lachapelle, Sara Magliacane, Perouz
  Taslakian, Georg Martius, Julius von K{\"u}gelgen, and Francesco Locatello.
\newblock Multi-view causal representation learning with partial observability.
\newblock \emph{arXiv preprint arXiv:2311.04056}, 2023.

\bibitem[Sch{\"o}lkopf et~al.(2021)Sch{\"o}lkopf, Locatello, Bauer, Ke,
  Kalchbrenner, Goyal, and Bengio]{scholkopf2021toward}
Bernhard Sch{\"o}lkopf, Francesco Locatello, Stefan Bauer, Nan~Rosemary Ke, Nal
  Kalchbrenner, Anirudh Goyal, and Yoshua Bengio.
\newblock Toward causal representation learning.
\newblock \emph{Proceedings of the IEEE}, 109\penalty0 (5):\penalty0 612--634,
  2021.

\bibitem[Liu et~al.(2023{\natexlab{b}})Liu, Alahi, Russell, Horn, Zietlow,
  Sch{\"o}lkopf, and Locatello]{Liu2023CausalTriplet}
Yuejiang Liu, Alexandre Alahi, Chris Russell, Max Horn, Dominik Zietlow,
  Bernhard Sch{\"o}lkopf, and Francesco Locatello.
\newblock Causal triplet: An open challenge for intervention-centric causal
  representation learning.
\newblock In \emph{2nd Conference on Causal Learning and Reasoning (CLeaR)},
  2023{\natexlab{b}}.

\bibitem[Gresele et~al.(2020)Gresele, Rubenstein, Mehrjou, Locatello, and
  Sch{\"o}lkopf]{gresele2020incomplete}
Luigi Gresele, Paul~K Rubenstein, Arash Mehrjou, Francesco Locatello, and
  Bernhard Sch{\"o}lkopf.
\newblock The incomplete rosetta stone problem: Identifiability results for
  multi-view nonlinear ica.
\newblock In \emph{Uncertainty in Artificial Intelligence}, pages 217--227.
  PMLR, 2020.

\bibitem[Comon(1994)]{comon1994independent}
Pierre Comon.
\newblock Independent component analysis, a new concept?
\newblock \emph{Signal processing}, 36\penalty0 (3):\penalty0 287--314, 1994.

\bibitem[Hyv{\"a}rinen(2013)]{hyvarinen2013independent}
Aapo Hyv{\"a}rinen.
\newblock Independent component analysis: recent advances.
\newblock \emph{Philosophical Transactions of the Royal Society A:
  Mathematical, Physical and Engineering Sciences}, 371\penalty0
  (1984):\penalty0 20110534, 2013.

\bibitem[Lee and Lee(1998)]{lee1998independent}
Te-Won Lee and Te-Won Lee.
\newblock \emph{Independent component analysis}.
\newblock Springer, 1998.

\bibitem[Zhang and Chan(2007)]{zhang2007kernel}
Kun Zhang and Laiwan Chan.
\newblock Kernel-based nonlinear independent component analysis.
\newblock In \emph{International Conference on Independent Component Analysis
  and Signal Separation}, pages 301--308. Springer, 2007.

\bibitem[Zheng et~al.(2022)Zheng, Ng, and Zhang]{zheng2022identifiability}
Yujia Zheng, Ignavier Ng, and Kun Zhang.
\newblock On the identifiability of nonlinear ica: Sparsity and beyond.
\newblock \emph{Advances in Neural Information Processing Systems},
  35:\penalty0 16411--16422, 2022.

\bibitem[Hyv{\"a}rinen and Pajunen(1999)]{hyvarinen1999nonlinear}
Aapo Hyv{\"a}rinen and Petteri Pajunen.
\newblock Nonlinear independent component analysis: Existence and uniqueness
  results.
\newblock \emph{Neural networks}, 12\penalty0 (3):\penalty0 429--439, 1999.

\bibitem[Hyv{\"a}rinen et~al.(2023)Hyv{\"a}rinen, Khemakhem, and
  Monti]{hyvarinen2023identifiability}
Aapo Hyv{\"a}rinen, Ilyes Khemakhem, and Ricardo Monti.
\newblock Identifiability of latent-variable and structural-equation models:
  from linear to nonlinear.
\newblock \emph{arXiv preprint arXiv:2302.02672}, 2023.

\bibitem[Khemakhem et~al.(2020{\natexlab{a}})Khemakhem, Monti, Kingma, and
  Hyvarinen]{khemakhem2020ice}
Ilyes Khemakhem, Ricardo Monti, Diederik Kingma, and Aapo Hyvarinen.
\newblock Ice-beem: Identifiable conditional energy-based deep models based on
  nonlinear ica.
\newblock \emph{Advances in Neural Information Processing Systems},
  33:\penalty0 12768--12778, 2020{\natexlab{a}}.

\bibitem[Li et~al.(2023{\natexlab{b}})Li, Xu, Cai, Yang, Yan, Hao, Chen, and
  Zhang]{li2023identifying}
Zijian Li, Zunhong Xu, Ruichu Cai, Zhenhui Yang, Yuguang Yan, Zhifeng Hao,
  Guangyi Chen, and Kun Zhang.
\newblock Identifying semantic component for robust molecular property
  prediction.
\newblock \emph{arXiv preprint arXiv:2311.04837}, 2023{\natexlab{b}}.

\bibitem[Khemakhem et~al.(2020{\natexlab{b}})Khemakhem, Kingma, Monti, and
  Hyvarinen]{khemakhem2020variational}
Ilyes Khemakhem, Diederik Kingma, Ricardo Monti, and Aapo Hyvarinen.
\newblock Variational autoencoders and nonlinear ica: A unifying framework.
\newblock In \emph{International Conference on Artificial Intelligence and
  Statistics}, pages 2207--2217. PMLR, 2020{\natexlab{b}}.

\bibitem[Hyvarinen and Morioka(2016)]{hyvarinen2016unsupervised}
Aapo Hyvarinen and Hiroshi Morioka.
\newblock Unsupervised feature extraction by time-contrastive learning and
  nonlinear ica.
\newblock \emph{Advances in neural information processing systems}, 29, 2016.

\bibitem[Hyvarinen and Morioka(2017)]{hyvarinen2017nonlinear}
Aapo Hyvarinen and Hiroshi Morioka.
\newblock Nonlinear ica of temporally dependent stationary sources.
\newblock In \emph{Artificial Intelligence and Statistics}, pages 460--469.
  PMLR, 2017.

\bibitem[Hyvarinen et~al.(2019)Hyvarinen, Sasaki, and
  Turner]{hyvarinen2019nonlinear}
Aapo Hyvarinen, Hiroaki Sasaki, and Richard Turner.
\newblock Nonlinear ica using auxiliary variables and generalized contrastive
  learning.
\newblock In \emph{The 22nd International Conference on Artificial Intelligence
  and Statistics}, pages 859--868. PMLR, 2019.

\bibitem[Xie et~al.(2022)Xie, Kong, Gong, and Zhang]{xie2022multi}
Shaoan Xie, Lingjing Kong, Mingming Gong, and Kun Zhang.
\newblock Multi-domain image generation and translation with identifiability
  guarantees.
\newblock In \emph{The Eleventh International Conference on Learning
  Representations}, 2022.

\bibitem[Kong et~al.(2023{\natexlab{b}})Kong, Huang, Xie, Xing, Chi, and
  Zhang]{kong2023identification}
Lingjing Kong, Biwei Huang, Feng Xie, Eric Xing, Yuejie Chi, and Kun Zhang.
\newblock Identification of nonlinear latent hierarchical models.
\newblock \emph{arXiv preprint arXiv:2306.07916}, 2023{\natexlab{b}}.

\bibitem[Yan et~al.(2023)Yan, Kong, Gui, Chi, Xing, He, and
  Zhang]{yan2023counterfactual}
Hanqi Yan, Lingjing Kong, Lin Gui, Yujie Chi, Eric Xing, Yulan He, and Kun
  Zhang.
\newblock Counterfactual generation with identifiability guarantees.
\newblock In \emph{37th International Conference on Neural Information
  Processing Systems, NeurIPS 2023}, 2023.

\bibitem[Lachapelle et~al.(2023)Lachapelle, Deleu, Mahajan, Mitliagkas, Bengio,
  Lacoste-Julien, and Bertrand]{lachapelle2023synergies}
S{\'e}bastien Lachapelle, Tristan Deleu, Divyat Mahajan, Ioannis Mitliagkas,
  Yoshua Bengio, Simon Lacoste-Julien, and Quentin Bertrand.
\newblock Synergies between disentanglement and sparsity: Generalization and
  identifiability in multi-task learning.
\newblock In \emph{International Conference on Machine Learning}, pages
  18171--18206. PMLR, 2023.

\bibitem[Lachapelle and Lacoste-Julien(2022)]{lachapelle2022partial}
S{\'e}bastien Lachapelle and Simon Lacoste-Julien.
\newblock Partial disentanglement via mechanism sparsity.
\newblock \emph{arXiv preprint arXiv:2207.07732}, 2022.

\bibitem[Zhang et~al.(2024{\natexlab{b}})Zhang, Xie, Ng, and
  Zheng]{zhang2024causal}
Kun Zhang, Shaoan Xie, Ignavier Ng, and Yujia Zheng.
\newblock Causal representation learning from multiple distributions: A general
  setting.
\newblock \emph{arXiv preprint arXiv:2402.05052}, 2024{\natexlab{b}}.

\bibitem[H{"a}lv{"a} and Hyvarinen(2020)]{halva2020hidden}
Hermanni H{"a}lv{"a} and Aapo Hyvarinen.
\newblock Hidden markov nonlinear ica: Unsupervised learning from nonstationary
  time series.
\newblock In \emph{Conference on Uncertainty in Artificial Intelligence}, pages
  939--948. PMLR, 2020.

\bibitem[Lippe et~al.(2022)Lippe, Magliacane, L{\"o}we, Asano, Cohen, and
  Gavves]{lippe2022citris}
Phillip Lippe, Sara Magliacane, Sindy L{\"o}we, Yuki~M Asano, Taco Cohen, and
  Stratis Gavves.
\newblock Citris: Causal identifiability from temporal intervened sequences.
\newblock In \emph{International Conference on Machine Learning}, pages
  13557--13603. PMLR, 2022.

\bibitem[Song et~al.(2023)Song, Yao, Fan, Dong, Chen, Niebles, Xing, and
  Zhang]{song2023temporally}
Xiangchen Song, Weiran Yao, Yewen Fan, Xinshuai Dong, Guangyi Chen, Juan~Carlos
  Niebles, Eric Xing, and Kun Zhang.
\newblock Temporally disentangled representation learning under unknown
  nonstationarity.
\newblock In \emph{Thirty-seventh Conference on Neural Information Processing
  Systems}, 2023.
\newblock URL \url{https://openreview.net/forum?id=V8GHCGYLkf}.

\bibitem[Daunhawer et~al.(2023)Daunhawer, Bizeul, Palumbo, Marx, and
  Vogt]{daunhawer2023identifiability}
Imant Daunhawer, Alice Bizeul, Emanuele Palumbo, Alexander Marx, and Julia~E
  Vogt.
\newblock Identifiability results for multimodal contrastive learning.
\newblock \emph{arXiv preprint arXiv:2303.09166}, 2023.

\end{thebibliography}
\bibliographystyle{unsrtnat}

\clearpage
\appendix
\newpage
\appendix
  \textit{\large Supplement to}\\ \ \\
      % {\Large \bf ``\texorpdfstring{\raisebox{-1mm}{\includegraphics[scale=0.45]{figures/icon.pdf}}}{NSCtrl}: Causal Temporal Representation Learning with Nonstationary Dynamics''}\
      {\Large \bf ``From Orthogonality to Dependency: Learning Disentangled Representation for Multi-Modal Time-Series Sensing Signals''}\
\newcommand{\beginsupplement}{%
	\setcounter{table}{0}
	\renewcommand{\thetable}{A\arabic{table}}%
	\setcounter{figure}{0}
	\renewcommand{\thefigure}{A\arabic{figure}}%
	\setcounter{section}{0}
	\renewcommand{\thesection}{A\arabic{section}}%
	\setcounter{theorem}{0}
	\renewcommand{\thetheorem}{A\arabic{theorem}}%
	\setcounter{corollary}{0}
	\renewcommand{\thecorollary}{A\arabic{corollary}}%	
}

\beginsupplement
{\large Appendix organization:}

\DoToC 
% \clearpage

% \clearpage
\section{Related Works}
\label{app:rl}
\subsection{Multi-modality Representation Learning}
Multimodality representation learning \cite{jiang2023understanding,liang2022high,tu2022cross,liu2023m3ae,radford2021learning} aims to mean information from different modalities, and have lots of applications like Visual Question Answering (VQA) \cite{duan2022multi,kwon2022masked,li2021align,shukor2022efficient,yang2022vision}. The mainstream methods include self-supervised learning \cite{jaiswal2020survey,zhai2019s4l,liu2021self}, masked autoencoders \cite{he2022masked,geng2022multimodal,kwon2022masked}, and the generative model-based methods \cite{limoyo2022learning,khattar2019mvae}. Multi-modality time series data is underexplored in literature, despite being often encountered in practice. One of the mainstream methods for multi-modality time series representation learning is to extract the modality-shared representation. Previously, Deldari et.al \cite{deldari2022cocoa} extracted the modality-shared representation by computing the cross-correlation of different modalities and minimizing the similarity between irrelevant instances. Deng \cite{deng2024multi} proposes multi-modality data augmentation to learn inter-modality and intra-modality representations. Recently, Kara \cite{kara2024freqmae} devised a factorized multi-modal fusion mechanism for leveraging cross-modal correlations to learn modality-specific representations. And Liu et.al \cite{liu2024focal} leverage both the modality-shared and modality-specific representation for downstream tasks. However, most of this method implicitly assumes that the latent space is orthogonal, which may be hard to meet in real-world scenarios. In this paper, we propose a data generation process with dependent subspace for mutli-modality time series data and devise a flexible model with theoretical guarantees.

\subsection{Identifiability of Generative Model}
To achieve identifiability \cite{rajendran2024learning,mansouri2023object,wendong2024causal} for causal representation, several researchers use the independent component analysis (ICA) to recover the latent variables with identification guarantees \cite{yao2023multi,scholkopf2021toward,Liu2023CausalTriplet,gresele2020incomplete}. Conventional methods assume a linear mixing function from the latent variables to the observed variables \cite{comon1994independent,hyvarinen2013independent,lee1998independent,zhang2007kernel}. Since the linear mixing process is hard to meet in real-world scenarios, recently, some researchers have established the identifiability via nonlinear ICA by using different types of assumptions like auxiliary variables or sparse generation process \cite{zheng2022identifiability,hyvarinen1999nonlinear,hyvarinen2023identifiability,khemakhem2020ice,li2023identifying}. Specifically, Aapo et.al \cite{khemakhem2020variational,hyvarinen2016unsupervised,hyvarinen2017nonlinear,hyvarinen2019nonlinear} first achieve the identifiability by assuming the latent sources with exponential family and introducing auxiliary variables e.g., domain indexes, time indexes, and class labels. And Zhang et.al \cite{kong2022partial, xie2022multi, kong2023identification,yan2023counterfactual} achieve the component-wise identification results for nonlinear ICA without using the exponential family assumption. To achieve identifiability without any supervised signals, several researchers employ sparsity assumptions \cite{zheng2022identifiability,hyvarinen1999nonlinear, hyvarinen2023identifiability,khemakhem2020ice,li2023identifying}. For example, Lachapelle et al. \cite{lachapelle2023synergies,lachapelle2022partial} introduced mechanism sparsity regularization as an inductive bias to identify causal latent factors. And Zhang et.al \cite{zhang2024causal} use the sparse structures of latent variables to achieve identifiability under distribution shift. Researchers also employ nonlinear ICA to achieve identifiability of time series data \cite{ yan2023counterfactual,huang2023latent,halva2020hidden, lippe2022citris}. For example, Aapo et.al \cite{hyvarinen2016unsupervised} ) adopt the independent sources premise and capitalize on the variability in variance across different data segments to achieve identifiability on nonstationary time series data. And Permutation-based contrastive learning is employed to identify the latent variables on stationary time series data. Recently, LEAP \cite{yao2021learning} and TDRL \cite{yao2022temporally} have adopted the properties of independent noises and variability historical information. And  Song et.al \cite{song2023temporally} identify latent variables without observed domain variables. 
As for the identifiability of modality, Imant et.al \cite{daunhawer2023identifiability} present the identifiability
results for multimodal contrastive learning. Yao et.al \cite{yao2023multi} consider the identifiability of multi-view causal representation under the partially observed settings. In this paper, we leverage the fairness of multi-modality data and variability historical information to achieve identifiability for multi-modality time series data.
\section{Proof of Modality-shared Latent Variables $\rvz^c_t$}
% \begin{figure}[h]
%     \centering
%     \includegraphics[width=0.5\textwidth]{figs/multimodality_generation.pdf}
%     \caption{Caption}
%     \label{fig:generation}
% \end{figure}
\subsection{Proof of Subspace Identification}
\label{app:a1}
\begin{theorem}
(\textbf{Subspace Identification of the Modality-shared and Modality-specific Latent Variables}) Suppose that the observed data from different modalities is generated following the data generation process in Figure \ref{fig:generation}, and we further make the following assumptions:
\begin{itemize}[leftmargin=*]
    \item A1 \underline{(Smooth and Positive Density:)} The probability density of latent variables is smooth and positive, i.e., $p(\rvz_t|\rvz_{t-1})>0$ over $\mathcal{Z}_t$ and $\mathcal{Z}_{t-1}$.
    \item A2 \underline{(Conditional Independence:)} Conditioned on $\rvz_{t-1}$, each $z_{t,i}^c$ is independent of $z_{t,j}^c$ for $i,j \in \{1,\cdots,n_c\}, i\neq j$. And conditioned on $\rvz_{t-1}$ and $\rvz_{t}^c$, each $z_{t,i}^{s_m}$ is independent of $z_{t,j}^{s_m}$, for $i,j \in \{n_c+1,\cdots,n\}, i\neq j$.
    % \item A3 \underline{(Independent Noise:)}
    \item A3 \underline{(non-singular Jacobian):} Each $g_m$ has non-singular Jacobian matrices almost anywhere and $g_m$ is invertible. 
    \item A4 \underline{(Linear Independence:)} 
    For any $z_t^{s_*} \in \mathcal{Z}_{t}^{s_*}$, there exist $n_{c}+1$ values of $\rvz_{t-1,k}^{s_m}, k = n_{c}+1,\cdots,n$, such that these vectors $\bm{v}_{t,j}$ are linearly independent, where $\bm{v}_{t,j}$ are defined as follows:
    \begin{equation}
    \small
    \begin{split}
        \bm{v}_{t,j}=\Big(\frac{\partial^2 \log p({z}_{t,j}^{s_m}|\rvz^m_{t-1},{\rvz}_{t}^c)}{\partial z_{t,j}^{s_m}\partial z_{t-1,n_c+1}^{s_m}},\cdots,\frac{\partial^2 \log p({z}_{t,j}^{s_m}|\rvz^m_{t-1},{\rvz}_{t}^c)}{\partial z_{t,j}^{s_m}\partial z_{t-1,n}^{s_m}}\Big)
    \end{split}
    \end{equation} 
\end{itemize}
Then if $\hat{g}_1:\mathcal{Z}_t^c\times\mathcal{Z}_t^{s_1}\rightarrow \mathcal{X}_t^{s_1}$ and $\hat{g}_2:\mathcal{Z}_t^c\times\mathcal{Z}_t^{s_2}\rightarrow \mathcal{X}_t^{s_2}$ assume the generating process of the true model $(g_1,g_2)$ and match the joint distribution $p(\rvx_t^{s_1}, \rvx_t^{s_2})$ of each time step then $\rvz^c_t$ is subspace identifiable.
\end{theorem}

\begin{proof}
    For $(\rvx_t^1, \rvx_t^2)\sim p(\rvx_t^1, \rvx_t^2)$, because of the matched joint distribution, we have the following relations between the true variables $\rvz_t^c, \rvz_t^{s_1},\rvz_t^{s_2}$ and the estimated ones $\hat{\rvz}_t^c, \hat{\rvz}_t^{s_1}, \hat{\rvz}_t^{s_2}$:
    \begin{equation}
    \label{the:1}
    \begin{split}
x_t^{s_1}&=g_1(\rvz_t^c,\rvz_t^{s_1})=\hat{g}_1(\hat{\rvz}_t^c,\hat{\rvz}_t^{s_1})
    \end{split}
    \end{equation}
    \begin{equation}
    \label{the:2}
x_t^{s_2}=g_2(\rvz_t^c,\rvz_t^{s_2})=\hat{g}_2(\hat{\rvz}_t^c,\hat{\rvz}_t^{s_2})
    \end{equation}
    \begin{equation}
    \label{the:3}
        (\hat{\rvz}_t^c, \hat{\rvz}_t^{s_1}, \hat{\rvz}_t^{s_2})=\hat{g}^{-1}(\rvx_t^{s_1}, \rvx_t^{s_2})=\hat{g}^{-1}(g(\rvz_t^c,\rvz_t^{s_1},\rvz_t^{s_2})):=h(\rvz_t^c,\rvz_t^{s_1},\rvz_t^{s_2}),
    \end{equation}
where $\hat{g}_1, \hat{g}_2$ are the estimated invertible generating function and $h:=\hat{g}^{-1}\circ g$ denotes a smooth and invertible function that transforms the true variables $\rvz_t^c, \rvz_t^{s_1},\rvz_t^{s_2}$ to the estimated ones $\hat{\rvz}_t^c, \hat{\rvz}_t^{s_1}, \hat{\rvz}_t^{s_2}$.

By combining Equation (\ref{the:3}) and (\ref{the:1}), we have 
\begin{equation}
\label{the:4}
    g_1(\rvz_t^c,\rvz_t^{s_1})=\hat{g}_1(h_{c,s_1}(\rvz_t^c,\rvz_t^{s_1},\rvz_t^{s_2})).
\end{equation}
For $i\in \{1, \cdots, n_{\rvx^{s_1}}\}$ and $j\in \{1, \cdots, n_{s_2}\}$, we take a partial derivative of the $i$-th dimension of $\rvx^{s_1}_t$ on both sides of Equation (\ref{the:4}) w.r.t. $\rvz_{t,j}^{s_2}$ and have:
\begin{equation}
    0=\frac{\partial g_{1,i}(\rvz_t^c,\rvz_{t}^{s_{1}})}{\partial z_{t,j}^{s_2}}=\frac{\partial \hat{g}_{1,i}(h_{c,s_1}(\rvz^c_t,\rvz^{s_1}_{t}))}{\partial z_{t,j}^{s_2}}.
\end{equation}
The aforementioned equation equals 0 because there is no $z_{t,j}^{s_2}$ in the left-hand side of the equation. By expanding the derivative on the right-hand side, we further have:
\begin{equation}
    \sum_{k\in\{1,\cdots,n_c+n_{s_1}\}}\frac{\partial \hat{g}_{1,i}(\rvz_t^c,\rvz_t^{s_1})}{\partial h_{(c,s_1),k}}\cdot\frac{\partial h_{(c,s_1),k}(\rvz_t^c,\rvz_t^{s_1},\rvz_t^{s_2})}{\partial z_{t,j}^{s_2}}=0.
\end{equation}
Since $\hat{g}_1$ is invertible, the determinant of $\mathrm{J}_{\hat{g}_1}$ does not equal to 0, meaning that for $n_c+n_{s_1}$ different values of $\hat{g}_{1,i}$, each vector $[\frac{\partial \hat{g}_{1,i}(\rvz_t^c,\rvz_t^{s_1})}{\partial h_{(c,s_1),1}},\cdots,\frac{\partial \hat{g}_{1,i}(\rvz_t^c,\rvz_t^{s_1})}{\partial h_{(c,s_1),n_c+n_{s_1}}}]$ are linearly independent. Therefore, the $(n_c+n_{s_1})\times (n_c+n_{s_1})$ linear system is invertible and has the unique solution as follows:
\begin{equation}
    \label{eg:s2}
    \frac{\partial h_{(c,s_1),k}(\rvz_t^c,\rvz_t^{s_1},\rvz_t^{s_2})}{\partial z_{t,j}^{s_2}}=0.
\end{equation}
According to Equation (\ref{eg:s2}), for any $k\in \{1,\cdots,n_c+n_{s_1}\}$ and $j\in\{1,\cdots,n_{s_2}\}$, $h_{(c,s_1),k}(\rvz_t^c,\rvz_t^{s_1},\rvz_t^{s_2})$ does not depend on $\rvz_t^{s_2}$. In other word, $\{\rvz_t^c,\rvz_t^{s_1}\}$ does not depend on $\rvz_t^{s_2}$.

Similarly, by combining Equation (\ref{the:3}) and (\ref{the:2}), we have
\begin{equation}
    \label{eg:s22}  g_2(\rvz_t^c,\rvz_t^{s_2})=\hat{g}_2(h_{c,s_2}(\rvz_t^c,\rvz_t^{s_1},\rvz_t^{s_2})).
\end{equation}
For $i\in\{1,\cdots,n_{\rvx^{s_2}}\}$ and $j\in\{1,\cdots,n_{s_1}\}$, we take a partial derivative of the $i$-th dimension of $\rvx^{s_2}_t$ on both sides of Equation (\ref{eg:s22}) w.r.t $z_{t,j}^{s_1}$ and have:
\begin{equation}
    0=\frac{\partial g_{2,i}(\rvz_t^c,\rvz_{t}^{s_{2}})}{\partial z_{t,j}^{s_1}}=\frac{\partial \hat{g}_{2,i}(h_{c,s_2}(\rvz^c_t,\rvz^{s_2}_{t})}{\partial z_{t,j}^{s_1}}= \sum_{k\in\{1\cdots,n_c+n_{s_2}\}}\frac{\partial \hat{g}_{2,i}(\rvz_t^c,\rvz_t^{s_2})}{\partial h_{(c,s_2),k}}\cdot\frac{\partial h_{(c,s_2),k}(\rvz_t^c,\rvz_t^{s_1},\rvz_t^{s_2})}{\partial z_{t,j}^{s_1}}
\end{equation}
Since $\hat{g}_2$ is invertible, for $n_c+n_{s_2}$ different values of $\hat{g}_{2,i}$, each vector $[\frac{\partial \hat{g}_{2,i}(\rvz_t^c,\rvz_t^{s_2})}{\partial h_{(c,s_2),1}},\cdots,\frac{\partial \hat{g}_{2,i}(\rvz_t^c,\rvz_t^{s_2})}{\partial h_{(c,s_2),n_c+n_{s_2}}}]$ are linearly independent. Therefore, the $(n_c+n_{s_2})\times(n_c+n_{s_2})$ linear system is invertible and has the unique solution as follows:
\begin{equation}
    \frac{\partial h_{(c,s_2),k}(\rvz_t^c,\rvz_t^{s_1},\rvz_t^{s_2})}{\partial z_{t,j}^{s_1}}=0,
\end{equation}
meaning that $\{\rvz_t^c,\rvz_t^{s_2}\}$ does not depend on $\rvz_t^{s_1}$.

According to Equation (\ref{the:3}), we have $\hat{\rvz}_t^c=h_c(\rvz_t^c,\rvz_t^{s_1},\rvz_t^{s_2})$. By using the fact that $\{\rvz_t^c,\rvz_t^{s_2}\}$ does not depend on $\rvz_t^{s_1}$ and $\{\rvz_t^c,\rvz_t^{s_1}\}$ does not depend on $\rvz_t^{s_2}$, we have $\hat{\rvz}_t^c=h_c(\rvz_t^c)$, i.e., the modality-shared latent variables are subspace identifiable.

Since the matched marginal distribution of $p(\rvx^{s_1}_t|\rvx^{s_1}_{t-1})$, we have:
\begin{equation}
    \forall \rvx^{s_1}_{t-1} \in \mathcal{X}^{s_1}_{t-1}, \quad p(\hat{\rvx}^{s_1}_t|\rvx^{s_1}_{t-1})=p(\rvx^{s_1}_t|\rvx^{s_1}_{t-1}) \Longleftrightarrow p(\hat{g}_1(\hat{\rvz}^1_t)|\rvx^{s_1}_{t-1})=p(g_1(\rvz^1_t)|\rvx^{s_1}_{t-1}),
\end{equation}
where $\rvz^1_t=\{\rvz_t^c,\rvz^{s_1}_t\}$ and $\hat{\rvz}^1_t=\{\hat{\rvz}_t^c,\hat{\rvz}^{s_1}_t\}$.
Sequentially, by using the change of variables formula, we can further obtain Equation (\ref{the5})
\begin{equation}
\begin{split}
\label{the5}
    p(\hat{g}_1(\hat{\rvz}^1_t)|\rvx^{s_1}_{t-1})=p(g_1(\rvz^1_t)|\rvx^{s_1}_{t-1}) &\Longleftrightarrow p(g^{-1}_1\circ \hat{g}_{1}(\hat{\rvz}^1_t)|\rvx^{s_1}_{t-1})|\mathbf{J}_{g_1^{-1}}|=p(\rvz^1_t|\rvx^{s_1}_{t-1})|\mathbf{J}_{g_1^{-1}}|\\&\Longleftrightarrow p(h_1(\hat{\rvz}^1_t)|\rvx^{s_1}_{t-1})=p(\rvz^1_t|\rvx^{s_1}_{t-1})\\&\Longleftrightarrow p(h_1(\hat{\rvz}^1_t)|\hat{\rvz}^1_{t-1})=p(\rvz^1_t|\rvz^1_{t-1}),
\end{split}
\end{equation}
where $h_1:=g^{-1}_1\circ \hat{g}_{1}$ is the transformation between the ground-true and the estimated latent variables. $\mathbf{J}_{g_1^{-1}}$ denotes the absolute value of Jacobian matrix determinant of $g_1^{-1}$. Since we assume that $g_1$ and $\hat{g}_1$ are invertible, $|\mathbf{J}_{g^{-1}}|\neq 0$ and $h_1$ is also invertible. 

According to the A2 (conditional independent assumption), we can have Equation (\ref{the6})
\begin{equation}
\label{the6}
    p(\rvz^1_t|\rvz^1_{t-1})=\prod_{i=1}^n p(z^1_{t,i}|\rvz^1_{t-1}); \quad p(\hat{\rvz}^1_t|\hat{\rvz}^1_{t-1})=\prod_{i=1}^n p(\hat{z}^1_{t,i}|\hat{\rvz}^1_{t-1}).
\end{equation}
For convenience, we take logarithm on both sides of Equation (\ref{the6}) and have:
\begin{equation}
\label{the7}
    \log p(\rvz^1_t|\rvz^1_{t-1}) = \sum_{i=1}^n \log p(z^1_{t,i}|\rvz^1_{t-1});\quad \log p(\hat{\rvz}^1_t|\hat{\rvz}^1_{t-1})=\sum_{i=1}^n \log p(\hat{z}^1_{t,i}|\hat{\rvz}^1_{t-1}).
\end{equation}
By combining Equation (\ref{the7}) and Equation (\ref{the5}), we have:
\begin{equation}
\begin{split}
    p(h_1(\hat{\rvz}^1_t)|\hat{\rvz}^1_{t-1})=p(\rvz^1_t|\rvz^1_{t-1}) &\Longleftrightarrow p(\hat{\rvz}^1_t|\hat{\rvz}^1_{t-1})|\mathbf{J}_{h^{-1}}|=p(\rvz^1_t|\rvz^1_{t-1})\\&\Longleftrightarrow \sum_{i=1}^n \log p(\hat{z}^1_{t,i}|\hat{\rvz}^1_{t-1})=\sum_{i=1}^n \log p(z^1_{t,i}|\rvz^1_{t-1}) - \log |\mathbf{J}_{h^{-1}}|,
\end{split}
\end{equation}
where $\mathbf{J}_{h^{-1}}$ are the Jacobian matrix of $h^{-1}$. 

Sequentially, we take the first-order derivative with $\hat{\rvz}^c_{t,i}$, where $i \in \{1,\cdots,n_c\}$ and have:
\begin{equation}
\begin{split}
    &\frac{\partial \log p(\hat{\rvz}^1_{t}|\hat{\rvz}^1_{t-1})}{\partial \hat{z}_{t,i}^c}=\sum_{j=1}^{n_c}\frac{\partial \log p(\hat{z}_{t,j}^c|\hat{\rvz}^1_{t-1})}{\partial \hat{z}_{t,i}^c} + \sum_{j=n_c+1}^n \frac{\partial \log p(\hat{z}_{t,j}^{s_1}|\hat{\rvz}^1_{t-1},\hat{\rvz}_{t}^c)}{\partial \hat{z}_{t,i}^c} \\=&\sum_{j=1}^{n_c}\frac{\partial \log p({z}_{t,j}^c|\rvz^1_{t-1})}{\partial z_{t,j}^c}\cdot\frac{\partial z_{t,j}^c}{\partial \hat{z}_{t,i}^c} + \sum_{j=n_c+1}^n \frac{\partial \log p({z}_{t,j}^{s_1}|\rvz^1_{t-1},{\rvz}_{t}^c)}{\partial z_{t,j}^{s_1}}\cdot\frac{\partial z_{t,j}^{s_1}}{\partial \hat{z}_{t,i}^c} - \frac{\partial |\mathbf{J}_{h^{-1}}|}{\partial \hat{z}_{t,i}^c}.
\end{split}
\end{equation}
Then we further take the second-order derivative w.r.t $\rvz_{t-1,k}^{s_1}$, where $k \in \{n_c+1,\cdots,n\}$ and we have:
\begin{equation}
\label{the8}
\begin{split}
    &\sum_{j=1}^{n_c}\frac{\partial^2 \log p(\hat{z}_{t,j}^c|\hat{\rvz}^1_{t-1})}{\partial \hat{z}_{t,i}^c \partial z_{t-1,k}^{s_1}} + \sum_{j=n_c+1}^n \frac{\partial^2 \log p(\hat{z}_{t,j}^{s_1}|\hat{\rvz}^1_{t-1},\hat{\rvz}_{t}^c)}{\partial \hat{z}_{t,i}^c \partial z_{t-1,k}^{s_1}} \\=&\sum_{j=1}^{n_c}\frac{\partial^2 \log p({z}_{t,j}^c|\rvz^1_{t-1})}{\partial z_{t,j}^c\partial z_{t-1,k}^{s_1}}\cdot\frac{\partial z_{t,j}^c}{\partial \hat{z}_{t,i}^c} + \sum_{j=n_c+1}^n \frac{\partial^2 \log p({z}_{t,j}^{s_1}|\rvz^1_{t-1},{\rvz}_{t}^c)}{\partial z_{t,j}^{s_1}\partial z_{t-1,k}^{s_1}}\cdot\frac{\partial z_{t,j}^{s_1}}{\partial \hat{z}_{t,i}^c} - \frac{\partial^2 |\mathbf{J}_{h^{-1}}|}{\partial \hat{z}_{t,i}^c\partial z_{t-1,k}^{s_1}}.
\end{split}
\end{equation}

Since $\hat{z}_{t,j}^c$ does not change across different values of $z_{t-1,k}^{s_1}$, then $\frac{\partial^2 \log p(\hat{z}_{t,j}^c|\hat{\rvz}^1_{t-1})}{\partial \hat{z}_{t,i}^c \partial z_{t-1,k}^{s_1}}=0$. Since $\frac{\partial^2 \log p(\hat{z}_{t,j}^{s_1}|\hat{\rvz}^1_{t-1},\hat{\rvz}_{t}^c)}{\partial \hat{z}_{t,i}^c}$ does not change across different values of $z_{t-1,k}^{s_1}$, then $\frac{\partial^2 \log p(\hat{z}_{t,j}^{s_1}|\hat{\rvz}^1_{t-1},\hat{\rvz}_{t}^c)}{\partial \hat{z}_{t,i}^c \partial z_{t-1,k}^{s_1}}=0$. Moreover, since $\frac{\partial^2 \log p({z}_{t,j}^c|\rvz^1_{t-1})}{\partial z_{t,j}^c\partial z_{t-1,k}^{s_1}}$ and $\frac{\partial^2 |\mathbf{J}_{h^{-1}}|}{\partial \hat{z}_{t,i}^c\partial z_{t-1,k}^{s_1}}=0$, Equation (\ref{the8}) can be further rewritten as:
\begin{equation}
\label{the9}
    \sum_{j=n_c+1}^n \frac{\partial^2 \log p({z}_{t,j}^{s_1}|\rvz^1_{t-1},{\rvz}_{t}^c)}{\partial z_{t,j}^{s_1}\partial z_{t-1,k}^{s_1}}\cdot\frac{\partial z_{t,j}^{s_1}}{\partial \hat{z}_{t,i}^c}=0.
\end{equation}
By leveraging the linear independence assumption, the linear system denoted by Equation (\ref{the9}) has the only solution $\frac{\partial z_{t,j}^{s_1}}{\partial \hat{z}_{t,i}^c}=0$. As $h_1$ is smooth, its Jacobian can written as:
\begin{equation}
\begin{gathered}
\label{eq:Jh1}
    \mJ_{h_1}=\begin{bmatrix}
    \begin{array}{c|c}
        \textbf{A}:=\frac{\partial \rvz_t^{c}}{\partial \hat{\rvz}_t^{c}} & \textbf{B}:=\frac{\partial \rvz_t^c}{\partial \hat{\rvz}_t^{s_1}}=0 \\ \midrule
        \textbf{C}:=\frac{\partial \rvz_t^{s_1}}{\partial \hat{\rvz}_t^c}=0 & \textbf{D}:=\frac{\partial \rvz_t^{s_1}}{\partial \hat{\rvz}_t^{s_1}}.
    \end{array}
    \end{bmatrix}
\end{gathered}
\end{equation}
Therefore, $\rvz_t^{s_1}$ is subspace identifiable. Similarly,we can prove that
$\rvz_t^{s_m}$ is subspace identifiable.
\end{proof}

\subsection{Proof of Component-wise Identification}
\label{app:a2}
\begin{corollary}
{(\textbf{Component-wise Identification of the Modality-shared and Modality-specific Latent Variables})} Suppose that the observed data from different modalities is generated following the data generation process in Figure \ref{fig:generation}, and we further make the following assumptions:
\begin{itemize}[leftmargin=*]
    \item A1 \underline{(Smooth and Positive Density:)} The probability density of latent variables is smooth and positive, i.e., $p(\rvz_t|\rvz_{t-1})>0$ over $\mathcal{Z}_t$ and $\mathcal{Z}_{t-1}$.
    \item A2 \underline{(Conditional Independence:)} Conditioned on $\rvz_{t-1}$, each $z_{t,i}^c$ is independent of $z_{t,j}^c$ for $i,j \in \{1,\cdots,n_c\}, i\neq j$. And conditioned on $\rvz_{t-1}$ and $\rvz_{t}^c$, each $z_{t,i}^{s_m}$ is independent of $z_{t,j}^{s_m}$, for $i,j \in \{n_c+1,\cdots,n\}, i\neq j$.
    % \item A3 \underline{(Independent Noise:)}
    % \item A3 \underline{(non-singular Jacobian):} Each $g_m$ has non-singular Jacobian matrices almost anywhere and $g_m$ is invertible. 
    \item A3 \underline{(Linear Independence:)} 
    For any $z_t \in \mathcal{Z}_{t}$, there exist $2n +1$ values of $\rvz^m_{t-1,k}, k =1,\cdots,n$, such that these vectors $\bm{v}_{t,l}$ are linearly independent, where $\bm{v}_{t,l}$ are defined as follows:
    \begin{equation}
    \small
    \begin{split}
        \bm{v}_{t,l}=\Big(\frac{\partial^3 \log p({z}_{t,l}^c|\rvz^m_{t-1})}{\partial^2 z_{t,l}^c\partial z^m_{t-1,1}},\cdots,\frac{\partial^3 \log p({z}_{t,l}^c|\rvz^m_{t-1})}{\partial^2 z_{t,l}^c\partial z^m_{t-1,n}},\frac{\partial^2 \log p({z}_{t,l}^c|\rvz^m_{t-1})}{\partial z_{t,l}^c\partial z^m_{t-1,1}},\cdots,\frac{\partial^2 \log p({z}_{t,l}^c|\rvz^m_{t-1})}{\partial z_{t,l}^c\partial z^m_{t-1,n}},
     \\\frac{\partial^3 \log p({z}_{t,l}^{s_m}|\rvz^m_{t-1},{\rvz}_{t}^c)}{\partial^2 z_{t,l}^{s_m}\partial z^m_{t-1,1}},\cdots,\frac{\partial^3 \log p({z}_{t,l}^{s_m}|\rvz^m_{t-1},{\rvz}_{t}^c)}{\partial^2 z_{t,l}^{s_m}\partial z^m_{t-1,n}},\frac{\partial^2 \log p({z}_{t,l}^{s_m}|\rvz^m_{t-1},{\rvz}_{t}^c)}{\partial z_{t,l}^{s_m}\partial z^m_{t-1,1}},\cdots,\frac{\partial^2 \log p({z}_{t,l}^{s_m}|\rvz^m_{t-1},{\rvz}_{t}^c)}{\partial z_{t,l}^{s_m}\partial z^m_{t-1,n}}\Big)
    \end{split}
    \end{equation} 
\end{itemize}
Then if $\hat{g}_1:\mathcal{Z}_t^c\times\mathcal{Z}_t^{s_1}\rightarrow \mathcal{X}_t^{s_1}$ and $\hat{g}_2:\mathcal{Z}_t^c\times\mathcal{Z}_t^{s_2}\rightarrow \mathcal{X}_t^{s_2}$ assume the generating process of the true model $(g_1,g_2)$ and match the joint distribution $p(\rvx_t^{s_1}, \rvx_t^{s_2})$ of each time step then $\rvz^c_t$ is component-wise identifiable.
\end{corollary}
\begin{proof}
Then we let $\rvz^1_t=\{\rvz_t^c,\rvz_t^{s_1}\}$ and $\hat{\rvz}^1_t=\{\hat{\rvz}_t^c,\hat{\rvz}_t^{s_1}\}$. According to Equation (2), we have $\hat{\rvz}_t=h_1(\rvz_t)$, where $h_1:=\hat{g_1}^{-1}\circ g_1$ is an invertible function. 
Sequentially, it is straightforward to see that if the components of $\hat{\rvz}_t^{s_1}$ are mutually independent conditional on $\hat{\rvz}_{t-1}^{s_1}$ and $\hat{\rvz}_{t}^{c}$, the components of $\hat{\rvz}_t^{c}$ are mutually independent conditional on $\hat{\rvz}_{t-1}^{c}$, then for any $i\neq j$, we have:
\begin{equation}
\label{eq:second derivative}
    \frac{\partial^2\log p(\hat{\rvz}_t^{s_1}|\hat{\rvz}_{t-1}^{s_1},\hat{\rvz}_t^{c})}{\partial\hat{z}_{t,i}^{s_1}\partial\hat{z}_{t,j}^{s_1}}=0,
    \frac{\partial^2\log p(\hat{\rvz}_t^{c}|\hat{\rvz}_{t-1}^{c})}{\partial\hat{z}_{t,i}^{c}\partial\hat{z}_{t,j}^{c}}=0,
\end{equation}
by assuming that the second-order derivative exists.
The Jacobian matrix of the mapping from $(\rvx_{t-1}^{s_1}, {\rvz}^1_t)$ to $(\rvx_{t-1}^{s_1},\hat{\rvz}^1_t)$ is 
$\small{
    \begin{bmatrix}
        \mathbb{I}&0\\
        * & H^{s_1}_t
\end{bmatrix}}$, where $H^{s_1}_t$ denotes the absolute value of the determinant of this Jacobian matrix is $|H^{s_1}_t|$. Therefore, $p(\hat{\rvz}^1_t,\rvx_{t-1}^{s_1})\cdot|H^{s_1}_t|=p(\rvz^1_t,\rvx_{t-1}^{s_1})$. Dividing both sides of this equation by $p(\rvx_{t-1}^{s_1})$ gives
\begin{equation}
    p(\hat{\rvz}^1_t|\rvx_{t-1}^{s_1})\cdot|H^{s_1}_t|=p(\rvz^1_t|\rvx_{t-1}^{s_1}).
\end{equation}
Since $p(\rvz^1_t|\rvz^1_{t-1})=p(\rvz^1_t|g_1(\rvz^1_{t-1}))=p(\rvz^1_t|\rvx_{t-1}^{s_1})$ and similarly $p(\hat{\rvz^1}_t|\hat{\rvz}^1_{t-1})=p(\hat{\rvz}^1_t|\rvx_{t-1}^{s_1})$, so we further have:
\begin{equation}\label{eq:var_change}
    \log p(\hat{\rvz}^1_t|\hat{\rvz}^1_{t-1})=\log p(\rvz^1_t|\rvz^1_{t-1})-\log|H_{t}^{s_1}|.
\end{equation}
According to Equation (\ref{eq:var_change}) , we take the first-order derivative with $\hat{\rvz}^c_{t,i}$, where $i \in \{1,\cdots,n_c\}$ and have:
\begin{equation}
\begin{split}
    &\frac{\partial \log p(\hat{\rvz}^1_{t}|\hat{\rvz}^1_{t-1})}{\partial \hat{z}_{t,i}^c}=\sum_{l=1}^{n_c}\frac{\partial \log p(\hat{z}_{t,l}^c|\hat{\rvz}^1_{t-1})}{\partial \hat{z}_{t,i}^c} + \sum_{l=n_c+1}^n \frac{\partial \log p(\hat{z}_{t,l}^{s_1}|\hat{\rvz}^1_{t-1},\hat{\rvz}_{t}^c)}{\partial \hat{z}_{t,i}^c} \\=&\sum_{l=1}^{n_c}\frac{\partial \log p({z}_{t,l}^c|\rvz^1_{t-1})}{\partial z_{t,l}^c}\cdot\frac{\partial z_{t,l}^c}{\partial \hat{z}_{t,i}^c} + \sum_{l=n_c+1}^n \frac{\partial \log p({z}_{t,l}^{s_1}|\rvz^1_{t-1},{\rvz}_{t}^c)}{\partial z_{t,l}^{s_1}}\cdot\frac{\partial z_{t,l}^{s_1}}{\partial \hat{z}_{t,i}^c} - \frac{\partial \log|H_{t}^{s_1}|}{\partial \hat{z}_{t,i}^c}.
\end{split}
\end{equation}
Then we further take the second-order derivative w.r.t $\hat\rvz_{t,j}^{c}$, where $j \in \{1,\cdots,n_c\}$ and we have:
\begin{equation}
\begin{split}
    &\sum_{l=1}^{n_c}\frac{\partial^2 \log p(\hat{z}_{t,l}^c|\hat{\rvz}^1_{t-1})}{\partial \hat{z}_{t,i}^c\partial \hat{z}_{t,j}^c} + \sum_{l=n_c+1}^n \frac{\partial^2 \log p(\hat{z}_{t,l}^{s_1}|\hat{\rvz}^1_{t-1},\hat{\rvz}_{t}^c)}{\partial \hat{z}_{t,i}^c\partial \hat{z}_{t,j}^c} \\=&\sum_{l=1}^{n_c}\frac{\partial^2 \log p({z}_{t,l}^c|\rvz^1_{t-1})}{\partial^2 z_{t,l}^c}\cdot\frac{\partial z_{t,l}^c}{\partial \hat{z}_{t,j}^c}\cdot\frac{\partial z_{t,l}^c}{\partial \hat{z}_{t,i}^c} +\sum_{l=1}^{n_c}\frac{\partial \log p({z}_{t,l}^c|\rvz^1_{t-1})}{\partial z_{t,l}^c}\cdot\frac{\partial^2 z_{t,l}^c}{\partial \hat{z}_{t,i}^c\partial \hat{z}_{t,j}^c} 
    \\ +&\sum_{l=n_c+1}^n \frac{\partial^2 \log p({z}_{t,l}^{s_1}|\rvz^1_{t-1},{\rvz}_{t}^c)}{\partial^2 z_{t,l}^{s_1}}\cdot\frac{\partial z_{t,l}^{s_1}}{\partial \hat{z}_{t,j}^c}\cdot\frac{\partial z_{t,l}^{s_1}}{\partial \hat{z}_{t,i}^c}+ \sum_{l=n_c+1}^n \frac{\partial \log p({z}_{t,l}^{s_1}|\rvz^1_{t-1},{\rvz}_{t}^c)}{\partial z_{t,l}^{s_1}}\cdot\frac{\partial^2 z_{t,l}^{s_1}}{\partial \hat{z}_{t,i}^c\partial \hat{z}_{t,j}^c} - \frac{\partial^2 \log|H_{t}^{s_1}|}{\partial \hat{z}_{t,i}^c\partial \hat{z}_{t,j}^c}.
\end{split}
\end{equation}
Sequentially, for $k=1,\cdots, n_c$, and each value $z_{t-1,k}^{c}$, the third-order derivative w.r.t. $v_{t-1,k}^{c}$, and we have:
\begin{equation}
\begin{split}
\label{eq:third-order}
    &\sum_{l=1}^{n_c}\frac{\partial^3 \log p(\hat{z}_{t,l}^c|\hat{\rvz}^1_{t-1})}{\partial \hat{z}_{t,i}^c\partial \hat{z}_{t,j}^c\partial {z}_{t-1,k}^c} + \sum_{l=n_c+1}^n \frac{\partial^3 \log p(\hat{z}_{t,l}^{s_1}|\hat{\rvz}^1_{t-1},\hat{\rvz}_{t}^c)}{\partial \hat{z}_{t,i}^c\partial \hat{z}_{t,j}^c\partial {z}_{t-1,k}^c} \\=&\sum_{l=1}^{n_c}\frac{\partial^3 \log p({z}_{t,l}^c|\rvz^1_{t-1})}{\partial^2 z_{t,l}^c\partial z_{t-1,k}^c}\cdot\frac{\partial z_{t,l}^c}{\partial \hat{z}_{t,j}^c}\cdot\frac{\partial z_{t,l}^c}{\partial \hat{z}_{t,i}^c} +\sum_{l=1}^{n_c}\frac{\partial^2 \log p({z}_{t,l}^c|\rvz^1_{t-1})}{\partial z_{t,l}^c\partial z_{t-1,k}^c}\cdot\frac{\partial^2 z_{t,l}^c}{\partial \hat{z}_{t,i}^c\partial \hat{z}_{t,j}^c} 
    \\ +&\sum_{l=n_c+1}^n \frac{\partial^3 \log p({z}_{t,l}^{s_1}|\rvz^1_{t-1},{\rvz}_{t}^c)}{\partial^2 z_{t,l}^{s_1}\partial z_{t-1,k}^c}\cdot\frac{\partial z_{t,l}^{s_1}}{\partial \hat{z}_{t,j}^c}\cdot\frac{\partial z_{t,l}^{s_1}}{\partial \hat{z}_{t,i}^c}+ \sum_{l=n_c+1}^n \frac{\partial^2 \log p({z}_{t,l}^{s_1}|\rvz^1_{t-1},{\rvz}_{t}^c)}{\partial z_{t,l}^{s_1}\partial z_{t-1,k}^c}\cdot\frac{\partial^2 z_{t,l}^{s_1}}{\partial \hat{z}_{t,i}^c\partial \hat{z}_{t,j}^c} - \frac{\partial^3 \log|H_{t}^{s_1}|}{\partial \hat{z}_{t,i}^c\partial \hat{z}_{t,j}^c\partial z_{t-1,k}^c}.
\end{split}
\end{equation}
Since according to Equation(\ref{eq:second derivative}),then $\frac{\partial^3 \log p(\hat{z}_{t,l}^c|\rvz^1_{t-1})}{\partial \hat{z}_{t,i}^c\partial \hat{z}_{t,j}^c\partial {z}_{t-1,k}^c}=0$.
 Since $\hat{z}_{t,l}^{s_1}$ does not change across different values of $z_{t-1,k}^{c}$, then $\frac{\partial^3 \log p(\hat{z}_{t,l}^{s_1}|\hat{\rvz}^1_{t-1},\hat{\rvz}_{t}^c)}{\partial \hat{z}_{t,i}^c\partial \hat{z}_{t,j}^c\partial {z}_{t-1,k}^c} = 0$.
 Equation (\ref{eq:third-order}) can be further rewritten as:
\begin{equation}
\begin{split}
\label{eq:finally eq}
    &\sum_{l=1}^{n_c}\frac{\partial^3 \log p({z}_{t,l}^c|\rvz^1_{t-1})}{\partial^2 z_{t,l}^c\partial z_{t-1,k}^c}\cdot\frac{\partial z_{t,l}^c}{\partial \hat{z}_{t,j}^c}\cdot\frac{\partial z_{t,l}^c}{\partial \hat{z}_{t,i}^c} +\sum_{l=1}^{n_c}\frac{\partial^2 \log p({z}_{t,l}^c|\rvz^1_{t-1})}{\partial z_{t,l}^c\partial z_{t-1,k}^c}\cdot\frac{\partial^2 z_{t,l}^c}{\partial \hat{z}_{t,i}^c\partial \hat{z}_{t,j}^c} 
    \\ +&\sum_{l=n_c+1}^n \frac{\partial^3 \log p({z}_{t,l}^{s_1}|\rvz^1_{t-1},{\rvz}_{t}^c)}{\partial^2 z_{t,l}^{s_1}\partial z_{t-1,k}^c}\cdot\frac{\partial z_{t,l}^{s_1}}{\partial \hat{z}_{t,j}^c}\cdot\frac{\partial z_{t,l}^{s_1}}{\partial \hat{z}_{t,i}^c}+ \sum_{l=n_c+1}^n \frac{\partial^2 \log p({z}_{t,l}^{s_1}|\rvz^1_{t-1},{\rvz}_{t}^c)}{\partial z_{t,l}^{s_1}\partial z_{t-1,k}^c}\cdot\frac{\partial^2 z_{t,l}^{s_1}}{\partial \hat{z}_{t,i}^c\partial \hat{z}_{t,j}^c} = 0.
\end{split}
\end{equation}
 where we have made use of the fact that entries of $H_t^{s_1}$ do not depend on $z_{t-1,l}^{c}$. Then by leveraging the linear independence assumption, the linear system denoted by Equation (\ref{eq:finally eq})  has the only solution $\frac{\partial z_{t,l}^{c}}{\partial \hat{z}_{t,i}^{c}}\cdot\frac{\partial z_{t,l}^{c}}{\partial \hat{z}_{t,j}^{c}}=0$ and $\frac{\partial^2 z_{t,l}^{c}}{\partial \hat{z}_{t,i}^{c}\partial \hat{z}_{t,j}^{c}}=0$ and $\frac{\partial z_{t,l}^{s_1}}{\partial \hat{z}_{t,i}^{c}}\cdot\frac{\partial z_{t,l}^{s_1}}{\partial \hat{z}_{t,j}^{c}}=0$ and $\frac{\partial^2 z_{t,l}^{s_1}}{\partial \hat{z}_{t,i}^{c}\partial \hat{z}_{t,j}^{c}}=0$. 
 According to Equation (\ref{eq:Jh1}),we have:
 \begin{equation}
\begin{gathered}
    \mJ_{h_1}=\begin{bmatrix}
    \begin{array}{c|c}
        \textbf{A}:=\frac{\partial \rvz_t^{c}}{\partial \hat{\rvz}_t^{c}} & \textbf{B}:=\frac{\partial \rvz_t^c}{\partial \hat{\rvz}_t^{s_1}}=0 \\ \midrule
        \textbf{C}:=\frac{\partial \rvz_t^{s_1}}{\partial \hat{\rvz}_t^c}=0 & \textbf{D}:=\frac{\partial \rvz_t^{s_1}}{\partial \hat{\rvz}_t^{s_1}}
    \end{array}
    \end{bmatrix}.
\end{gathered}
\end{equation}
Since $h_1$ is invertible and for $i,j\in \{1,\cdots,n_c\}$, $\frac{\partial z_{t,l}^{c}}{\partial \hat{z}_{t,i}^{c}}\cdot\frac{\partial z_{t,l}^{c}}{\partial \hat{z}_{t,j}^{c}}=0$ and $\frac{\partial z_{t,l}^{s_1}}{\partial \hat{z}_{t,i}^{c}}\cdot\frac{\partial z_{t,l}^{s_1}}{\partial \hat{z}_{t,j}^{c}}=0$ implies that for each $k=1,\cdots,n_c$, there is exactly one non-zero component in each column of matrices \textbf{A} and \textbf{C}. Since we have proved that $\hat{\rvz}_t^c=h_c(\rvz_t^c)$ and \textbf{C} = 0, there is exactly one non-zero component in each column of matrices \textbf{A}. Therefore, $\rvz_t^c$ is component-wise identifiable.

Based on Equation(\ref{eq:var_change}), we further let $i,j,k\in \{n_c+1,\cdots,n\}$, and its three-order derivation w.r.t. $\hat{z}_{t,i}^{s_1},\hat{z}_{t,j}^{s_1},z_{t-1,l}^{s_1}$ can be written as 
\begin{equation}
\begin{split}
    &\sum_{l=1}^{n_c}\frac{\partial^3 \log p({z}_{t,l}^c|\rvz^1_{t-1})}{\partial^2 z_{t,l}^c\partial z_{t-1,k}^{s_1}}\cdot\frac{\partial z_{t,l}^c}{\partial \hat{z}_{t,j}^{s_1}}\cdot\frac{\partial z_{t,l}^c}{\partial \hat{z}_{t,i}^{s_1}} +\sum_{l=1}^{n_c}\frac{\partial^2 \log p({z}_{t,l}^c|\rvz^1_{t-1})}{\partial z_{t,l}^c\partial z_{t-1,k}^{s_1}}\cdot\frac{\partial^2 z_{t,l}^c}{\partial \hat{z}_{t,i}^{s_1}\partial \hat{z}_{t,j}^{s_1}} 
    \\ +&\sum_{l=n_c+1}^n \frac{\partial^3 \log p({z}_{t,l}^{s_1}|\rvz^1_{t-1},{\rvz}_{t}^c)}{\partial^2 z_{t,l}^{s_1}\partial z_{t-1,k}^{s_1}}\cdot\frac{\partial z_{t,l}^{s_1}}{\partial \hat{z}_{t,j}^{s_1}}\cdot\frac{\partial z_{t,l}^{s_1}}{\partial \hat{z}_{t,i}^{s_1}}+ \sum_{l=n_c+1}^n \frac{\partial^2 \log p({z}_{t,l}^{s_1}|\rvz^1_{t-1},{\rvz}_{t}^c)}{\partial z_{t,l}^{s_1}\partial z_{t-1,k}^{s_1}}\cdot\frac{\partial^2 z_{t,l}^{s_1}}{\partial \hat{z}_{t,i}^{s_1}\partial \hat{z}_{t,j}^{s_1}} = 0.
\end{split}
\end{equation}
By using the linear independence assumption, the linear system denoted by Equation (\ref{eq:finally eq}) has the only solution $\frac{\partial z_{t,l}^{c}}{\partial \hat{z}_{t,i}^{s_1}}\cdot\frac{\partial z_{t,l}^{c}}{\partial \hat{z}_{t,j}^{s_1}}=0$ and $\frac{\partial^2 z_{t,l}^{c}}{\partial \hat{z}_{t,i}^{s_1}\partial \hat{z}_{t,j}^{s_1}}=0$ and $\frac{\partial z_{t,l}^{s_1}}{\partial \hat{z}_{t,i}^{s_1}}\cdot\frac{\partial z_{t,l}^{s_1}}{\partial \hat{z}_{t,j}^{s_1}}=0$ and $\frac{\partial^2 z_{t,l}^{s_1}}{\partial \hat{z}_{t,i}^{s_1}\partial \hat{z}_{t,j}^{s_1}}=0$,
meaning that there is exactly one non-zero component in each row of $\textbf{B}$ and $\textbf{D}$. Since $\textbf{B}=0$ ,then $\rvz_t^{s_1}$ is component-wise identifiable. Similarly, we can prove that $\rvz_t^{s_m}$ is component-wise identifiable. 

\end{proof}
\section{Evidence Lower Bound}\label{app:elbo}
% \label{app:a3}
In this subsection, we show the evidence lower bound. We first factorize the conditional distribution according to the Bayes theorem.

\begin{equation}
\tiny
\begin{split}
&\ln p(\rvx^{s_1}_{1:T},\rvx^{s_2}_{1:T})=\ln\frac{p(\rvx^{s_1}_{1:T},\rvx^{s_2}_{1:T},\rvz^{s_1}_{1:T},\rvz^{s_2}_{1:T},\rvz^{c}_{1:T})}{p(\rvz^{s_1}_{1:T},\rvz^{s_2}_{1:T},\rvz^{c}_{1:T}|\rvx^{s_1}_{1:T},\rvx^{s_2}_{1:T})}
% \\\\=&
% \ln \frac{p(\rvx_{t+1:T}|\rvz_{t+1:T}^e,\rvz_{t+1:T}^s)p(\rvx_{1:t}|\rvz_{1:t}^e,\rvz_{1:t}^s)p(\rvz^s_{1:t})p(\rvz_{1:t}^e)p(\rvz^s_{t+1:T}|\rvz^s_{1:t})p(\rvz^e_{t+1:T}|\rvz^e_{1:t})}{p(\rvz^e_{1:t}|\rvx_{1:T})p(\rvz^e_{t+1:T}|\rvx_{1:T},\rvz^e_{1:t})p(\rvz^s_{1:t}|\rvx_{1:T},\rvz^e_{1:T})p(\rvz^s_{1:T}|\rvx_{1:T},\rvz^e_{1:T},\rvz^s_{1:t})}
\\\\=&\ln\frac{p(\rvx^{s_1}_{1:T}|\rvz^{s_1}_{1:T},\rvz^{c}_{1:T})p(\rvx^{s_2}_{1:T}|\rvz^{s_2}_{1:T},\rvz^{c}_{1:T})p(\rvz^{s_1}_{1:T}|\rvz^{c}_{1:T})p(\rvz^{s_2}_{1:T}|\rvz^{c}_{1:T})p(\rvz^{c}_{1:T})}{p(\rvz^{s_1}_{1:T}|\rvx^{s_1}_{1:T},\rvz^{c}_{1:T})p(\rvz^{s_2}_{1:T}|\rvx^{s_2}_{1:T},\rvz^{c}_{1:T}))
p(\rvz^{c}_{1:T}|\rvx^{s_1}_{1:T},\rvx^{s_2}_{1:T})}
\\\\=&\mathbb{E}_{q(\rvz^{s_1}_{1:T}|\rvx^{s_1}_{1:T},\rvz^{c}_{1:T})}
\mathbb{E}_{q(\rvz^{s_2}_{1:T}|\rvx^{s_2}_{1:T},\rvz^{c}_{1:T})}
\mathbb{E}_{q(\rvz^c_{1:T}|\rvx^{s_1}_{1:T},\rvx^{s_2}_{1:T})}
\ln\frac{p(\rvx^{s_1}_{1:T}|\rvz^{s_1}_{1:T},\rvz^{c}_{1:T})p(\rvx^{s_2}_{1:T}|\rvz^{s_2}_{1:T},\rvz^{c}_{1:T})p(\rvz^{c}_{1:T})p(\rvz^{s_1}_{1:T}|\rvz^{c}_{1:T})p(\rvz^{s_2}_{1:T}|\rvz^{c}_{1:T})}{q(\rvz^{s_1}_{1:T}|\rvx^{s_1}_{1:T},\rvz^{c}_{1:T})q(\rvz^{s_2}_{1:T}|\rvx^{s_2}_{1:T},\rvz^{c}_{1:T})
q(\rvz^{c}_{1:T}|\rvx^{s_1}_{1:T},\rvx^{s_2}_{1:T})}
\\\\&+ D_{KL}(q(\rvz^{s_1}_{1:T}|\rvx^{s_1}_{1:T},\rvz^{c}_{1:T})||p(\rvz^{s_1}_{1:T}|\rvx^{s_1}_{1:T},\rvz^{c}_{1:T}))+ D_{KL}(q(\rvz^{s_2}_{1:T}|\rvx^{s_2}_{1:T},\rvz^{c}_{1:T})||p(\rvz^{s_2}_{1:T}|\rvx^{s_2}_{1:T},\rvz^{c}_{1:T}))
\\\\&+ D_{KL}(q(\rvz^{c}_{1:T}|\rvx^{s_1}_{1:T},\rvx^{s_2}_{1:T})||p(\rvz^{c}_{1:T}|\rvx^{s_1}_{1:T},\rvx^{s_2}_{1:T})) 
\\\\\geq&\mathbb{E}_{q(\rvz^{s_1}_{1:T}|\rvx^{s_1}_{1:T},\rvz^{c}_{1:T})}
\mathbb{E}_{q(\rvz^{s_2}_{1:T}|\rvx^{s_2}_{1:T},\rvz^{c}_{1:T})}
\mathbb{E}_{q(\rvz^c_{1:T}|\rvx^{s_1}_{1:T},\rvx^{s_2}_{1:T})}\ln \frac{p(\rvx^{s_1}_{1:T}|\rvz^{s_1}_{1:T},\rvz^{c}_{1:T})p(\rvx^{s_2}_{1:T}|\rvz^{s_2}_{1:T},\rvz^{c}_{1:T})p(\rvz^{c}_{1:T})p(\rvz^{s_1}_{1:T}|\rvz^{c}_{1:T})p(\rvz^{s_2}_{1:T}|\rvz^{c}_{1:T})}{q(\rvz^{s_1}_{1:T}|\rvx^{s_1}_{1:T},\rvz^{c}_{1:T})q(\rvz^{s_2}_{1:T}|\rvx^{s_2}_{1:T},\rvz^{c}_{T})
q(\rvz^{c}_{1:T}|\rvx^{s_1}_{1:T},\rvx^{s_2}_{1:T})}
\\\\=&
\underbrace{
\mathbb{E}_{q(\rvz^{s_1}_{1:T}|\rvx^{s_1}_{1:T},\rvz^{c}_{1:T})}
\mathbb{E}_{q(\rvz^{s_2}_{1:T}|\rvx^{s_2}_{1:T},\rvz^{c}_{1:T})}
\mathbb{E}_{q(\rvz^c_{1:T}|\rvx^{s_1}_{1:T},\rvx^{s_2}_{1:T})}
 \ln p(\rvx^{s_1}_{1:T}|\rvz^{s_1}_{1:T},\rvz^{c}_{1:T})p(\rvx^{s_2}_{1:T}|\rvz^{s_2}_{1:T},\rvz^{c}_{1:T})}_{\mathcal{L}_{rec}}
\\&- \underbrace{D_{KL}(q(\rvz^{s_1}_{1:T}|\rvx^{s_1}_{1:T},\rvz^{c}_{1:T})||p(\rvz^{s_1}_{1:T}|\rvz^{c}_{1:T}))}_{\mathcal{L}_{s_1}}- \underbrace{D_{KL}(q(\rvz^{s_1}_{1:T}|\rvx^{s_2}_{1:T},\rvz^{c}_{1:T})||p(\rvz^{s_2}_{1:T}|\rvz^{c}_{1:T}))}_{\mathcal{L}_{s_2}}- \underbrace{D_{KL}(q(\rvz^{c}_{1:T}|\rvx^{s_1}_{1:T},\rvx^{s_2}_{1:T})||p(\rvz^{c}_{1:T}))}_{\mathcal{L}_{c}}
\end{split}
% \vspace{-6mm}
\end{equation}

\section{Prior Estimation}
\label{app:a4}
\textbf{Shared Prior Estimation: }We first consider the prior of $\ln p(\rvz^{c}_{1:T} )$. We consider the time lag as $L=1$,we devise a transformation $\sigma^c:= \{\hat{\rvz}_{t-1}^c, \hat{\rvz}_t^c\}\rightarrow\{\hat{\rvz}_{t-1}^c, \hat{\epsilon}_{t}^c\}$. Then we write this latent process as a transformation map $\mathbf{\sigma}$ (note that we overload the notation $\sigma$ for transition functions and for the transformation map):
    \begin{equation}
    % \small
\begin{gathered}\nonumber
    \begin{bmatrix}
    \begin{array}{c}
        \hat{z}^{c}_{t-1}  \\ 
        \hat{z}^{c}_{t}  \\
    \end{array}
    \end{bmatrix}=\mathbf{\sigma}\left(
    \begin{bmatrix}
    \begin{array}{c}
        \hat{z}^{c}_{t-1}  \\ 
   \hat{\epsilon}^{c}_{t}   
    \end{array}
    \end{bmatrix}\right).
\end{gathered}
\end{equation}
By applying the change of variables formula to the map $\mathbf{f}$, we can evaluate the joint distribution of the latent variables $p(\hat{z}^{c}_{t-1} 
        \hat{z}^{c}_{t} )$ as 
\begin{equation}
\small
\label{equ:p1}
    p(\hat{z}^{c}_{t-1}, 
        \hat{z}^{c}_{t} )=\frac{p(\hat{z}^{c}_{t-1}  \\ 
   \hat{\epsilon}^{c}_{t} )}{|\text{det }\mathbf{J}_{\mathbf{\sigma}}|},
\end{equation}
where $\mathbf{\sigma}_{\mathbf{\sigma}}$ is the Jacobian matrix of the map $\mathbf{f}$, which is naturally a low-triangular matrix:
\begin{equation}
\small
\begin{gathered}\nonumber
\mathbf{J}_{\mathbf{\sigma}}=\begin{bmatrix}
    \begin{array}{cc}
        1 & 0  \\
        \frac{\partial \hat{z}^{c}_{t} }{\partial \hat{z}^{c}_{t-1} } & \frac{\hat{z}^{c}_{t}}{\hat{\epsilon}^{c}_{t}}
    \end{array}
    \end{bmatrix}.
\end{gathered}
\end{equation}

 Let $\{r^{c}_i \}_{i=1,2,3,\cdots}$ be a set of learned inverse transition functions that take the estimated latent causal variables, and output the noise terms, i.e., $\hat{\epsilon}_{t,i} =r^{c}_i (\hat{z}^{c}_{t,i} ,  \hat{\rvz}^{c}_{t-1} )$. Then we design a transformation $\mathbf{A}\rightarrow \mathbf{B}$ with low-triangular Jacobian as follows:
\begin{equation}
\small
\begin{gathered}
    \label{eq:ssss}\underbrace{[\hat{\rvz}^{c}_{t-1} ,{\hat{\rvz}^{c}}_{t} ]^{\top}}_{\mathbf{A}} \text{  mapped to  } \underbrace{[{\hat{\rvz}^{c}}_{t-1},{\hat{\epsilon}^{c}}_{t} ]^{\top}}_{\mathbf{B}}, \text{ with } \mathbf{J}_{\mathbf{A}\rightarrow\mathbf{B}}=
    \begin{bmatrix}
        \mathbb{I}&0\\
        * & \text{diag}\left(\frac{\partial r^c_i}{\partial \hat{z}_{t-1,i}^c}\right)
    \end{bmatrix}.
\end{gathered}
\end{equation}
Similar to Equation (\ref{eq:ssss}), we can obtain the joint distribution of the estimated dynamics subspace as:
\begin{equation}
\small
    \log p(\mathbf{A})=\log p(\mathbf{B}) +\log (|\text{det}(\mathbf{J}_{\mathbf{A}\rightarrow\mathbf{B}})|).
\end{equation}
Finally, we have:
\begin{equation}
\small
\label{equ:prior_c22}
\log p(\hat{\rvz}^c_{t}|\rvz^c_{t-1})=\log p(\hat{\epsilon}^c_{t}) + \sum_{i=n_d+1}^n\log|\frac{\partial r_i^c}{\partial \hat{z}^c_{t-1,i}}|.
\end{equation}
As a result, the prior distribution shared latent variables can be estimated as follows:
\begin{equation}
 \small
\label{equ:prior_c33}
p(\hat{\rvz}_{1:T}^c)=p(\hat{\rvz}^c_1)\prod_{\tau=2}^T \left( \sum_{i=n_d+1}^n\log p(\hat{\epsilon}^c_{\tau,i}) +\sum_{i=n_d+1}^n \log|\frac{\partial r_i^c}{\partial \hat{z}^c_{\tau-1,i}}| \right),
\end{equation}
where we assume $p(\hat{\epsilon}^c_{\tau,i})$ follows a standard Gaussian distribution.\\

As for the modality-specific prior estimation, we can obtain a similar derivation, by considering the modality-shared prior as condition.

\section{Implementation Details}
\label{app:a5}
We summarize our network architecture below and describe it in detail in Table A1. We also provide the training details on Table A2 and A3.

% \vspace{-12cm}
\begin{table}[h]
\renewcommand{\arraystretch}{0.65}
\small
\centering
% \vspace{-3cm}
\caption{Architecture details. BS: batch size, T: length of time series, LeakyReLU: Leaky Rectified Linear Unit, $|\rvx_t|$: the dimension of $\rvx_t$.}
\label{tab:architecture}
\begin{tabular}{@{}c|cc@{}}
\toprule
\textbf{Configuration}       & \multicolumn{1}{c|}{\textbf{Description}}                            & \textbf{Output}         \\ 
\midrule
1. $\psi_c$                    & \multicolumn{1}{c|}{Modality-shared Encoder}              &                         \\ \midrule
Input:$x_{1:T}$              & \multicolumn{1}{c|}{Observed time series}                            & BS $\times t \times |\rvx_T|$  \\
Augmentations                      & \multicolumn{1}{c|}{Time-Domain Transpose}                                & BS $\times 2$ $\times t$ $\times$ $|\rvx_T|$   \\
CNN Block                        & \multicolumn{1}{c|}{150 neurons}                           & BS $\times$ $t$ $\times 150$   \\
CNN Block                        & \multicolumn{1}{c|}{150 neurons}                           & BS $\times$ $t$ $\times 150$   \\
Permute                        & \multicolumn{1}{c|}{Matrix Transpose}                                       & BS $\times$  $ 150$ $\times t$   \\
GRU                      & \multicolumn{1}{c|}{300 neurons}                                & BS $\times 300 $    \\
Split                        & \multicolumn{1}{c|}{Transpose}                                       & BS $\times$  $ t$$\times n_c$ 
\\ 
 \midrule

2. $\psi_s$                    & \multicolumn{1}{c|}{Modality-private Encoder}              &                         \\ \midrule
Input:$x_{1:T}$              & \multicolumn{1}{c|}{Observed time series}                            & BS $\times t \times |\rvx_T|$  \\
Augmentations                      & \multicolumn{1}{c|}{Time-Domain Transpose}                                & BS $\times 2$ $\times t$ $\times$ $|\rvx_T|$   \\
CNN Block                        & \multicolumn{1}{c|}{150 neurons}                           & BS $\times$ $t$ $\times 150$   \\
CNN Block                        & \multicolumn{1}{c|}{150 neurons}                           & BS $\times$ $t$ $\times 150$   \\
Permute                        & \multicolumn{1}{c|}{Matrix Transpose}                                       & BS $\times$  $ 150$ $\times t$   \\
GRU                      & \multicolumn{1}{c|}{300 neurons}                                & BS $\times 300 $    \\

Split                        & \multicolumn{1}{c|}{Transpose}                                       & BS $\times$  $ t $$\times n_c$ 
\\
Dense                        & \multicolumn{1}{c|}{ $n_s$ neurons}                                     & BS $\times$  $ t $$\times n_s$ 
\\
 \midrule

3.$F_x$                        & \multicolumn{1}{c|}{Reconstruction Decoder}                              &                         \\ \midrule
Input:$z^c_{1:T},z^s_{1:T}$ & \multicolumn{1}{c|}{Modality-share and Modality-privte Latent Variable} & BS$\times$ t $\times n_c$, BS$\times$ t$\times n_s$ \\
Concat                       & \multicolumn{1}{c|}{concatenation}                                   & BS $\times$ t $\times$ ($n_c+n_s$)   \\
Dense                        & \multicolumn{1}{c|}{x dimension neurons}                                   & BS $\times$ t $\times |x_T|$          \\

 \midrule
4.$F_y$                        & \multicolumn{1}{c|}{Downstream task Predictor}                                                       &                         \\ \midrule
Input:$z^c_{1:T},z^s_{1:T}$ & \multicolumn{1}{c|}{Modality-share and Modality-private Latent Variable} & BS $\times t$ $\times n_c$ ,BS$\times t$ $\times n_s$ \\
Concat                       & \multicolumn{1}{c|}{concatenation}                                   & BS $\times t$ $\times$ ($n_s+n_c$)   \\
Dense                        & \multicolumn{1}{c|}{x neurons,GELU}                           & BS $\times t$ $\times x$           \\
Dense                        & \multicolumn{1}{c|}{n neurons}                                     & BS  $\times t$ $\times n$            \\
\midrule
{ 5.$r_c$} & \multicolumn{1}{c|}{Modality-share Prior Networks}                       &                         \\ \midrule
Input:$z^c_{1:T}$               & \multicolumn{1}{c|}{Latent Variables}                  & BS $\times$ $(n_c+1)$       \\
Dense                        & \multicolumn{1}{c|}{128 neurons,LeakyReLU}                           & $(n_c+1)\times 128$             \\
Dense                        & \multicolumn{1}{c|}{128 neurons,LeakyReLU}                           & 128$\times$128                 \\
Dense                        & \multicolumn{1}{c|}{128 neurons,LeakyReLU}                           & 128$\times$128                 \\
Dense                        & \multicolumn{1}{c|}{1 neuron}                                        & BS $\times$1                \\
JacobianCompute              & \multicolumn{1}{c|}{Compute log ( abs(det (J)))}                          & BS              \\  \midrule
{ 6.$r_s$} & \multicolumn{1}{c|}{Modality-private Prior Networks}                       &                         \\ \midrule
Input:$z^s_{1:T}$ and $z^c_{1:T}$               & \multicolumn{1}{c|}{Latent Variables}                  & BS $\times$ $(n_c+n_s+1)$       \\
Dense                        & \multicolumn{1}{c|}{128 neurons,LeakyReLU}                           & $(n_c+n_s+1)\times 128$             \\
Dense                        & \multicolumn{1}{c|}{128 neurons,LeakyReLU}                           & 128$\times$128                 \\
Dense                        & \multicolumn{1}{c|}{128 neurons,LeakyReLU}                           & 128$\times$128                 \\
Dense                        & \multicolumn{1}{c|}{1 neuron}                                        & BS $\times$1                \\
JacobianCompute              & \multicolumn{1}{c|}{Compute log (abs( det (J)))}                          & BS              \\ 
\bottomrule
\end{tabular}%
\end{table}

\section{Experiment Details}
\label{app:a6}

% \vspace{-12cm}
\begin{table}[h]
\renewcommand{\arraystretch}{0.95}
\small
\centering
% \vspace{-3cm}
\caption{Supervised Training Congfigurations(We use LR for Learning Rate).}
\label{tab:architecture}
\begin{tabular}{c|cc|cc|cc|cc|cc|cc|cc|}
\toprule
    \textbf{Dataset}
    & \multicolumn{1}{c|}{Motion}   
    & \multicolumn{1}{c|}{DINAMO} 
    & \multicolumn{1}{c|}{WIFI} 
    & \multicolumn{1}{c}{KETI} 
    & \multicolumn{1}{c}{HumanEVA} 
    & \multicolumn{1}{c}{H36M} 
    & \multicolumn{1}{c}{MIT-BIH} 
 \\ 
\midrule
Temperature             & 
\multicolumn{1}{c|}{0.5}      
&\multicolumn{1}{c|}{0.5}
&\multicolumn{1}{c|}{0.5}
&\multicolumn{1}{c|}{0.5} 
&\multicolumn{1}{c|}{0.5}  
&\multicolumn{1}{c|}{0.5}  
&\multicolumn{1}{c}{0.5}  
\\
Batch Size                      & 
\multicolumn{1}{c|}{32}      & \multicolumn{1}{c|}{64}
&\multicolumn{1}{c|}{32}
&\multicolumn{1}{c|}{64}
&\multicolumn{1}{c|}{64}
&\multicolumn{1}{c|}{64}
&\multicolumn{1}{c}{64}  \\
Window Length                        & 
\multicolumn{1}{c|}{256}      & \multicolumn{1}{c|}{256}
&\multicolumn{1}{c|}{256}
&\multicolumn{1}{c|}{256}
&\multicolumn{1}{c|}{75}
&\multicolumn{1}{c|}{125}
&\multicolumn{1}{c}{64}  \\
Supervised Optimizer                        & \multicolumn{1}{c|}{AdawW}      & \multicolumn{1}{c|}{AdawW}
&\multicolumn{1}{c|}{AdawW}
&\multicolumn{1}{c|}{AdawW}
&\multicolumn{1}{c|}{AdawW}
&\multicolumn{1}{c|}{AdawW}
&\multicolumn{1}{c}{AdawW}  \\
Supervised Max LR                       & 
\multicolumn{1}{c|}{1e-4}      & \multicolumn{1}{c|}{1e-4}
&\multicolumn{1}{c|}{1e-4}
&\multicolumn{1}{c|}{1e-4}
&\multicolumn{1}{c|}{1e-4}
&\multicolumn{1}{c|}{1e-4}
&\multicolumn{1}{c}{1e-4}  \\
Supervised Min LR                      & 
\multicolumn{1}{c|}{1e-6}      & \multicolumn{1}{c|}{1e-6}
&\multicolumn{1}{c|}{1e-6}
&\multicolumn{1}{c|}{1e-6}
&\multicolumn{1}{c|}{1e-6}
&\multicolumn{1}{c|}{1e-6}
&\multicolumn{1}{c}{1e-6}  \\
Supervised Scheduler                        & \multicolumn{1}{c|}{cosine}   &\multicolumn{1}{c|}{cosine}
&\multicolumn{1}{c|}{cosine}
&\multicolumn{1}{c|}{cosine}
&\multicolumn{1}{c|}{cosine}
&\multicolumn{1}{c|}{cosine}
&\multicolumn{1}{c}{cosine}  \\

 \bottomrule
\end{tabular}%
\end{table}
\begin{table}[h]
\renewcommand{\arraystretch}{0.95}
\small
\centering
% \vspace{-3cm}
\caption{Self-Supervised Training Congfigurations(We use LR for Learning Rate).}
\label{tab:architecture}
\begin{tabular}{c|cc|cc|cc|}
\toprule
    \textbf{Dataset}
    & \multicolumn{1}{c|}{UCIHAR}   
    & \multicolumn{1}{c|}{RealWorld-HAR} 
    & \multicolumn{1}{c}{PAMAP2} 

 \\ 
\midrule
Temperature             & 
\multicolumn{1}{c|}{0.5}      & \multicolumn{1}{c|}{0.5}
&\multicolumn{1}{c}{0.5}  \\
Batch Size                      & 
\multicolumn{1}{c|}{64}      & \multicolumn{1}{c|}{64}
&\multicolumn{1}{c}{64}  \\
Window Length                        & 
\multicolumn{1}{c|}{128}      & \multicolumn{1}{c|}{150}
&\multicolumn{1}{c}{512}  \\
Supervised Optimizer                        & \multicolumn{1}{c|}{AdawW}      & \multicolumn{1}{c|}{--}
&\multicolumn{1}{c}{--}  \\
Supervised Max LR                       & 
\multicolumn{1}{c|}{1e-4}  
&\multicolumn{1}{c|}{--}
&\multicolumn{1}{c}{--}  \\
Supervised Min LR                      & 
\multicolumn{1}{c|}{1e-6}     
&\multicolumn{1}{c|}{--}
&\multicolumn{1}{c}{--}  \\
Supervised Scheduler                        & \multicolumn{1}{c|}{cosine}   
&\multicolumn{1}{c|}{--}
&\multicolumn{1}{c}{--}  \\
Pretrain Optimizer                        & 
\multicolumn{1}{c|}{AdawW}   
&\multicolumn{1}{c|}{AdawW}
&\multicolumn{1}{c}{AdawW}  \\
Pretrain Max LR                       & 
\multicolumn{1}{c|}{1e-3}      & \multicolumn{1}{c|}{1e-3}

&\multicolumn{1}{c}{1e-3}  \\
Pretrain Min LR                      & 
\multicolumn{1}{c|}{1e-7}      & \multicolumn{1}{c|}{1e-7}

&\multicolumn{1}{c}{1e-7}  \\
Pretrain Scheduler                        & 
\multicolumn{1}{c|}{cosine}   &\multicolumn{1}{c|}{cosine}

&\multicolumn{1}{c}{cosine}  \\
Pretrain Weight Decay                        & \multicolumn{1}{c|}{0.5}      & \multicolumn{1}{c|}{0.5}
&\multicolumn{1}{c}{0.5}  \\
Finetune Optimizer                        & 
\multicolumn{1}{c|}{AdawW}      & \multicolumn{1}{c|}{AdawW}
&\multicolumn{1}{c}{--}  \\
Finetune Start LR                        & 
\multicolumn{1}{c|}{1e-3}      & \multicolumn{1}{c|}{1e-3}
&\multicolumn{1}{c}{--}  \\
Finetune Scheduler                        & 
\multicolumn{1}{c|}{cosine}   &\multicolumn{1}{c|}{cosine}
&\multicolumn{1}{c}{--}  \\
Finetune LR Decay                        & 
\multicolumn{1}{c|}{0.2}      & \multicolumn{1}{c|}{0.2}
&\multicolumn{1}{c}{--}  \\
Finetune LR Period                       & 
\multicolumn{1}{c|}{50}      & \multicolumn{1}{c|}{50}
&\multicolumn{1}{c}{--}  \\
Finetune Epochs                       & 
\multicolumn{1}{c|}{200}      & \multicolumn{1}{c|}{200}
&\multicolumn{1}{c}{--}  \\
 \bottomrule
\end{tabular}%
\end{table}
\subsection{Dataset Descriptions}
\label{ap:a61}
\begin{table}[h]
\renewcommand{\arraystretch}{0.95}
\small
\centering
% \vspace{-3cm}
\caption{Statistical Summaries of Evaluated Datasets.}
\label{tab:architecture}
\begin{tabular}{c|cc|cc|cc|}
\toprule
    \textbf{Dataset}
    & \multicolumn{1}{c|}{Modalities}   
    & \multicolumn{1}{c|}{Windows} 
    & \multicolumn{1}{c}{Classes} 

 \\ 
\midrule
Motion             & 
\multicolumn{1}{c|}{5x(Acc,Gyro,Mag)
(back, left L-arm, right
U-arm, left/right shoe)}      & \multicolumn{1}{c|}{256}
&\multicolumn{1}{c}{4}  \\
D1NAMO                      & 
\multicolumn{1}{c|}{ECG(lead II, lead V1)}      & \multicolumn{1}{c|}{256}
&\multicolumn{1}{c}{2}  \\
WIFI                        & 
\multicolumn{1}{c|}{Wireless x3}      & \multicolumn{1}{c|}{256}
&\multicolumn{1}{c}{7}  \\
KETI                        & \multicolumn{1}{c|}{4 sensors (monitoring CO2, temperature, humidity and light intensity)}      & \multicolumn{1}{c|}{256}
&\multicolumn{1}{c}{2}  \\
HumanEVA                       & 
\multicolumn{1}{c|}{Skeleton x15}  
&\multicolumn{1}{c|}{75}
&\multicolumn{1}{c}{5}  \\
H36M                      & 
\multicolumn{1}{c|}{Skeleton x17}     
&\multicolumn{1}{c|}{125}
&\multicolumn{1}{c}{15}  \\
UCIHAR                         & \multicolumn{1}{c|}{Body ACC,Total ACC, Total Gyro}   
&\multicolumn{1}{c|}{128}
&\multicolumn{1}{c}{6}  \\
MIT-BIH                         & \multicolumn{1}{c|}{Heart Rate, Breathing Rate, Avg Acceleration, Peak Acceleration}   
&\multicolumn{1}{c|}{64}
&\multicolumn{1}{c}{5}  \\
Relative-World                        & 
\multicolumn{1}{c|}{acc, gyro, mag}   
&\multicolumn{1}{c|}{150}
&\multicolumn{1}{c}{8}  \\
PAMAP2                      & 
\multicolumn{1}{c|}{acc, gyro}      & \multicolumn{1}{c|}{512}

&\multicolumn{1}{c}{18}  \\

 \bottomrule
\end{tabular}%
\end{table}
In this paper, we consider the WIFI \cite{yousefi2017survey}, and KETI \cite{hong2017high} datasets. Moreover, we further consider the human motion prediction datasets like Motion \cite{roggen2010collecting}, HumanEva-I \cite{sigal2010humaneva}, H36M \cite{ionescu2013human3}, UCIHAR \cite{anguita2013public}, PAMAP2 \cite{reiss2012introducing}, and RealWorld-HAR \cite{sztyler2016body}which consider different positions of the human body as different modalities. Moreover, we also consider two healthcare datasets such as MIT-BIH \cite{moody2001impact} and D1NAMO \cite{dubosson2018open}, which are related to arrhythmia and noninvasive type 1 diabetes. 

\textbf{Motion \cite{roggen2010collecting}} dataset is a subset of the OPPORTUNITY Activity Recognition Dataset \cite{roggen2010collecting}. Following the experimental setting of a recent device-based HAR study \cite{jeyakumar2019sensehar}, we consider 5 sensors worn at 5 different locations on the human body: left lower arm, left upper arm, right lower arm, right upper arm and the back. Each device contains an accelerometer, a gyroscope, and a magnetometer, and all three sensors generate three-axis readings. We focus on a 4-class prediction consisting of high-level locomotion activities (sit, stand, walk and lie).

\textbf{D1NAMO} \cite{dubosson2018open} is acquired on 20 healthy subjects and 9 patients with type-1 diabetes. The acquisition has been made in real-life conditions with the Zephyr BioHarness 3 wearable device. The dataset consists of ECG, breathing, and accelerometer signals, as well as glucose measurements and annotated food pictures. 

\textbf{WIFI} \cite{yousefi2017survey} dataset contains the amplitude and phase of wireless signals sent by three antennas. Each antenna transmits at 30 subcarriers, and the receiver base sampling frequency is 1000 Hz. The dataset contains 7 classes of activity, including lying down, falling, picking up, running, sitting down, standing up and walking. We also use a sliding window of 256 timestamps to get the segmented examples.

\textbf{KETI \cite{hong2017high}} dataset was collected from 51 rooms in a large university office building. Each room is instrumented with 4 sensors monitoring CO2, temperature, humidity and light intensity, with occupancy monitored by an additional PIR sensor in the room. Readings are recorded every 10 seconds, and the dataset contains one week worth of data. In this experiment, we target at human occupation prediction using the readings of these sensors.

\textbf{HumanEVA-I \cite{sigal2010humaneva}} comprises 3 subjects each performing 5 actions. We apply the original frame rate (60 Hz) and a 15-joint skeleton removing the root joint to build human motions. 

\textbf{H36M \cite{ionescu2013human3}} consists of 7 subjects (S1, S5, S6, S7, S8 ,S9 and S11) performing 15 different motions. We apply the original frame rate (50 Hz) and a 17-joint skeleton removing the root joint to build human motions. 

\textbf{UCHIHAR} \cite{anguita2013public} dataset contains recordings from 30 volunteers who carried out 6 classes of activities, including walking, walking upstairs, walking downstairs, sitting, standing, and lying. Activities are recorded by a smartphone device mounted on the volunteer’s waist.

% intro from https://nrl.northumbria.ac.uk/id/eprint/44509/1/Transactions_DNN_paper_2_5_2020_1_EH%20%281%29.pdf. please check
\textbf{MIT-BIH} \cite{moody2001impact} contains 48 records obtained from 47 subjects. Each subject is represented by one ECG recording using two leads: lead II (MLII) and lead V1. The sampling frequency of the signal is 360 Hz. The upper signal is lead II (MLII) and the lower signal is lead V1, obtained by placing
the electrodes on the chest. In the upper signal, the normal QRS complexes are usually prominent.

\textbf{RealWorld-HAR} \cite{sztyler2016body} is a public dataset using an accelerometer, gyroscope, magnetometer, and light signals from the forearm, thigh, head, upper arm, waist, chest, and shin to recognize eight common human activities performed by 15 subjects, including climbing stairs down and up, jumping, lying, standing, sitting, running/jogging, and walking.In our experiments, we only used the data collected from the “shin” sensor, including the accelerometer (ACC) and gyroscope. The sampling rate for all selected sensors was set at 100Hz.

\textbf{PAMAP2}  \cite{reiss2012introducing}  contains data on 18 different classes of physical activities performed by 9 subjects wearing 3 inertial measurement units and a heart rate monitor. In this set of experiments, we only used 3 accelerometer sensor data and 18 activities. Only data collected from the "chest" is used in our experiment

\begin{table*}[t]
\centering
\caption{Time series classification for Motion, Seizure, WIFI, and KETI datasets.}
\renewcommand{\arraystretch}{1.0}
\label{tab:app_cls1}
\resizebox{\textwidth}{!}{
\begin{tabular}{c|cc|cc|cc|cc}
\toprule
    & \multicolumn{2}{c|}{Motion}   
    & \multicolumn{2}{c|}{DINAMO} 
    & \multicolumn{2}{c|}{WIFI} 
    & \multicolumn{2}{c}{KETI} \\ \midrule
\textbf{Model}    & Accuracy        & Macro-F1  & Accuracy      & Macro-F1       & Accuracy            & Macro-F1
&Accuracy  &Macro-F1\\   
    \midrule
    \textbf{ResNet}        &
    89.96(0.234)&	91.41(0.139)&	88.64(0.262)&	88.58(0.273)&	90.29(0.519)&	88.14(0.648)&	96.05(0.387)&	
    84.59(1.181)\\
    \textbf{MaCNN}        & 
    85.57(2.117)&	86.93(2.429)&	90.17(0.172)&	48.56(1.666)&	88.81(3.821)&	87.80(3.353)&	93.05(1.411)&	71.93(2.178)
\\
    \textbf{SenenHAR}        & 
    88.95(0.369)&	88.66(0.276)&	89.56(0.620)&	47.23(0.182)&	94.63(0.614)&	92.75(0.686)&	96.43(0.143)&	84.74(0.379)
 \\
    \textbf{STFNets}        & 
    89.07(0.098)&	88.84(0.229)&	90.51(0.450)&	47.50(0.132)&	80.52(0.245)&	75.93(1.262)&	89.21(0.808)&	69.55(0.476)
 \\
    \textbf{RFNet-base}        & 
    89.93(0.281)&	91.70(0.408)&	90.76(0.252)&	58.79(4.911)&	86.31(1.765)&	82.56(2.313)&	95.12(0.478)&	81.45(1.077)
 \\
    \textbf{THAT}        & 
    89.66(0.488)&	91.38(0.521)&	92.76(0.292)&	71.64(2.229)&	95.59(1.027)&	94.86(1.126)&	96.33(0.283)&	85.12(1.143)
 \\
    \textbf{LaxCat}        &  
    60.25(3.678)&	41.01(4.381)&	90.64(0.362)&	54.56(2.013)&	76.36(1.492)&	73.85(2.155)&	93.33(1.449)&	70.67(0.335)
\\
    \textbf{UniTS}        & 
    91.02(0.399)&	92.73(0.432)&	90.88(0.362)&	58.39(4.048)&	95.83(0.812)&	94.49(1.383)&	96.04(0.613)&	84.08(1.601)
 \\
    \textbf{COCOA}        & 
    88.31(0.254)&	89.27(0.702)&	90.69(0.189)&	55.00(1.495)&	87.76(0.531)&	84.51(0.728)&	92.68(1.062)&	74.72(1.987)
\\
    \textbf{FOCAL}        &
    89.37(0.083)&	90.91(0.191)&	90.52(0.220)&	52.00(2.104)&	94.15(0.208)&	92.68(0.377)&	94.88(0.371)&	78.47(1.043)
 \\
    \textbf{CroSSL}        & 
    91.32(0.992)&	89.94(1.353)&	91.05(0.438)&	53.13(0.781)&	76.80(2.206)&	68.45(3.054)&	93.63(0.504)&	76.25(1.538)
 \\
    \midrule
    \textbf{MATED}        &  
    \textbf{92.44(0.160)}&	
    \textbf{93.75(0.154)}&	
    \textbf{93.31(0.170)}&	
    \textbf{73.72(1.148)}&
    \textbf{96.95(0.231)}&	
    \textbf{96.20(0.431)}&	
    \textbf{97.00(0.097)}&	
    \textbf{86.93(0.924)}
 \\

    \bottomrule
    \end{tabular}%
  \label{tab:exp_class_1}%
}
\end{table*}%
\begin{table*}[t]
\centering
\caption{Time series classification for human motion prediction and healthcare datasets.}
\renewcommand{\arraystretch}{1.0}
\label{tab:app_cls2}
\resizebox{\textwidth}{!}{
\begin{tabular}{c|cc|cc|cc|cc}
\toprule
    & \multicolumn{2}{c|}{HumanEVA} 
    & \multicolumn{2}{c|}{H36M} 
    & \multicolumn{2}{c|}{UCIHAR} 
    & \multicolumn{2}{c}{MIT-BIH} 
 \\ \midrule
\textbf{Model}    & Accuracy        & Macro-F1  & Accuracy      & Macro-F1       & Accuracy            & Macro-F1
&Accuracy  &Macro-F1\\   
    \midrule
        \textbf{ResNet}        &
        86.68(0.327)&	86.51(0.247)&	92.44(0.278)&	92.27(0.289)&	93.12(0.630)&	93.01(0.637)&	98.52(0.066)&	97.62(0.083)	
\\
    \textbf{MaCNN}        & 
    86.27(0.047)&	86.12(0.041)&	78.54(0.430)&	77.73(0.647)&	84.57(0.851)&	84.06(0.936)&	97.26(0.186)&	96.07(0.194)	
\\
    \textbf{SenenHAR}        & 
    85.77(1.078)&	86.00(1.185)&	67.69(0.525)&	67.44(0.490)&	87.77(1.228)&	87.47(1.252)&	95.82(0.036)&	94.79(0.735)	
 \\
    \textbf{STFNets}        & 
    86.07(0.368)&	85.76(0.291)&	61.67(1.481)&	57.20(1.112)&	81.64(0.521)&	81.64(0.339)&	91.63(0.369)&	88.97(0.217)	
 \\
    \textbf{RFNet-base}        & 
    97.15(0.616)&	96.18(0.457)&	94.14(0.674)&	93.14(0.710)&	95.63(0.952)&	95.16(1.414)&	98.64(0.139)&	97.85(0.108)	
 \\
    \textbf{THAT}        & 
    85.95(0.226)&	85.90(0.207)&	81.28(0.351)&	81.27(0.182)&	93.06(0.364)&	93.06(0.422)&	98.49(0.159)&	97.56(0.237)	
 \\
    \textbf{LaxCat}        &  
    86.28(0.023)&	86.20(0.045)&	86.09(2.516)&	85.84(2.495)&	89.00(0.476)&	88.78(0.429)&	97.77(0.113)&	96.77(0.131)	
\\
    \textbf{UniTS}        & 
    97.90(0.561)&	97.52(0.879)&	94.96(0.461)&	94.81(0.152)&	94.75(0.526)&	94.72(0.528)&	98.75(0.078)&	97.95(0.099)	
 \\
    \textbf{COCOA}        & 
    93.46(0.293)&	91.63(1.469)&	84.12(1.670)&	83.85(1.820)&	94.11(0.425)&	93.96(0.616)&	97.76(0.241)&	96.64(0.979)	
\\
    \textbf{FOCAL}        &
    92.15(1.428)&	91.83(1.214)&	89.73(0.270)&	89.30(0.282)&	94.36(0.098)&	94.36(0.190)&	98.67(0.053)&	97.84(0.103)	
 \\
    \textbf{CroSSL}        & 
   86.29(0.045)&	86.06(0.273)&	87.35(1.447)&	83.62(1.546)&	94.45(0.170)&	93.83(0.530)&	97.96(0.167)&	95.06(0.071)	
 \\
    \midrule
    \textbf{MATED}        &  
\textbf{98.90(0.108)}&	\textbf{98.82(0.094)}&	\textbf{96.12(0.036)}&	\textbf{95.99(0.037)}&	\textbf{95.97(0.258)}&	\textbf{95.93(0.273)}&	\textbf{98.97(0.065)}&	\textbf{98.34(0.147)}	
 \\

    \bottomrule
    \end{tabular}%
  \label{tab:exp_class_2}%
}
\end{table*}%
\subsection{More Experiment Results}
\label{app:a6}

\subsubsection{Ablation Studies}
\begin{figure}[H]
    \centering
    \includegraphics[width=1.0\columnwidth]{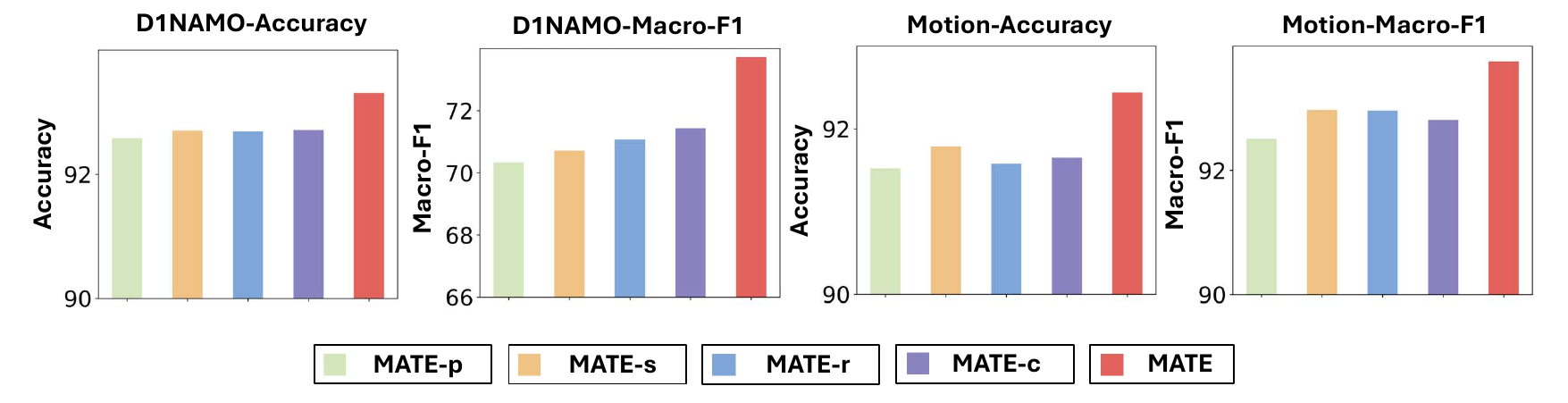}
    % \vspace{-2mm}
    \caption{Ablation study on the DINAMO and the Motion datasets. 
    % We explore the effectiveness of each loss term of the proposed method.
    }
    \label{fig:ablation}
\end{figure}

\begin{figure}[h]
	\centering
	\includegraphics[width=0.95\columnwidth]{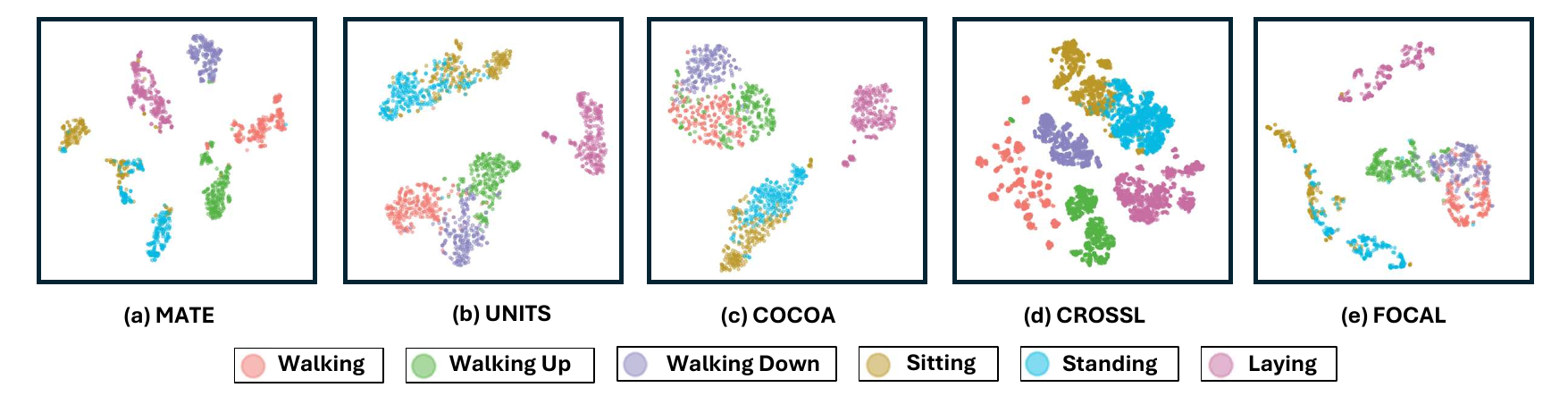}
	% \vspace{-3mm}
	\caption{The t-SNE visualization of the extracted domain-shared latent variables. }
	\label{fig:tsne}
\end{figure}

Figure \ref{fig:ablation} provides the results of ablation studies.

\subsubsection{Full Experiment Results}\label{app:more_exp}
Table A5 and Table A6 show the full results for the classification task.

\section{Limitation}\label{app:limitation}
Although our method can learn disentangled representation for multi-modal time series data with identifiability guarantees, it requires the assumption that the mixing function is invertible. However, this assumption might be hard to meet in real-world scenarios. Therefore, how to leverage the temporal context information to address this challenge will be an interesting direction. 

\section{Broader Impacts}\label{app:broader}
The proposed \textbf{MATE} model extracts the disentangled modality-shared and modality-specific latent variables for multi-modal time series modeling, which benefits the construction of 
precise and robust systems for time series data.

% \clearpage
% \bibliography{main}
% \bibliographystyle{plain}
% \end{document}
% \clearpage
% \subfile{paper_check_list.tex}
% \input{supplementary}

% \section{Appendix}

% Optionally include extra information (complete proofs, additional experiments and plots) in the appendix.
% This section will often be part of the supplemental material.

\end{document}